\documentclass[times,onecolumn,final,longtitle]{elsarticle}

\usepackage{smhl}
\usepackage{framed,multirow}
\usepackage{amssymb}
\usepackage{latexsym}
\usepackage{booktabs,siunitx,caption}
\usepackage{multicol}
\usepackage{csquotes}
\usepackage{adjustbox}
\usepackage{subcaption}
\usepackage{tablefootnote}
\usepackage{threeparttable, tablefootnote}	
\usepackage{color, colortbl}
\usepackage{graphicx}
\usepackage{url}
\usepackage{xcolor,soul}
\usepackage{hyperref}
\journal{Smart Health}

\begin{document}

\verso{Given-name Surname \textit{etal}}

\begin{frontmatter}

\title{Explainable AI for Malnutrition Risk Prediction from m-Health and Clinical Data}
\author[1]{Flavio \snm{Di Martino}\corref{cor1}}
\ead{flavio.dimartino@iit.cnr.it}
\author[1]{Franca \snm{Delmastro}}
\ead{franca.delmastro@iit.cnr.it}
\author[2]{Cristina \snm{Dolciotti}}
\ead{cristina.dolciotti@gmail.com}

\address[1]{Institute for Informatics and Telematics (IIT), National Research Council of Italy (CNR), Via Moruzzi 1, Pisa, 56100, Italy}
\address[2]{Dept. of Translational Research of New Technologies in Medicine and Surgery, University of Pisa, Pisa, 56126, Italy}

\received{1 May 2013}
\finalform{10 May 2013}
\accepted{13 May 2013}
\availableonline{15 May 2013}
\communicated{S. Sarkar}

\begin{abstract} 
Malnutrition is a serious and prevalent health problem in the older population, and especially in hospitalised or institutionalised subjects.
Accurate and early risk detection is essential for malnutrition management and prevention.
M-health services empowered with Artificial Intelligence (AI) may lead to important improvements in terms of a more automatic, objective, and continuous monitoring and assessment.
Moreover, the latest Explainable AI (XAI) methodologies may make AI decisions interpretable and trustworthy for end users.
This paper presents a novel AI framework for early and explainable malnutrition risk detection based on heterogeneous m-health data. 
We performed an extensive model evaluation including both subject-independent and personalised predictions, and the obtained results indicate Random Forest (RF) and Gradient Boosting as the best performing classifiers, especially when incorporating body composition assessment data.
We also investigated several benchmark XAI methods to extract global model explanations. Model-specific explanation consistency assessment indicates that each selected model privileges similar subsets of the most relevant predictors, with the highest agreement shown between SHapley Additive ExPlanations (SHAP) and feature permutation method.
Furthermore, we performed a preliminary clinical validation to verify that the learned feature-output trends are compliant with the current evidence-based assessment.
\end{abstract}
\end{frontmatter}

\section{Introduction}\label{sec:intro}
Malnutrition is a major public health issue, which is predominant in specific categories of frail subjects, such as children in developing countries \citep{collins2006management} and hospitalised/institutionalised older adults \citep{jensen2010adult}.
Healthcare institutions are in current need of solutions to support early and automatic assessment of individuals at higher risk conditions, in order to timely trigger proper feedback and interventions.
Although digitisation and Artificial Intelligence (AI) are used in many clinical applications, nutritional assessment and counseling still heavily rely on retrospective, analog questionnaires or consensus guidelines \citep{cederholm2019glim}, with different screening tools designed for specific population categories.
The Mini Nutritional Assessment (MNA) questionnaire and its short-form (MNA-SF) \citep{valentini2018frailty} currently represent the gold standard for geriatric healthcare professionals, since they include an evaluation of cognitive impairments, dementia, and depression in the overall assessment, whereas other tools are tailored to oncology patients \citep{li2019prognostic,martin2010prognostic}, as well as intensive care unit patients \citep{jeong2018comparison}.
These tools are administered by medical specialists by mostly interviewing the subject and/or the caregiver, suffering from subjective and recall biases. Moreover, they are usually collected during spot-checks, so they do not enable a continuous monitoring.
\\
The recent spread of smart technological solutions in the domain of human nutrition, especially in terms of m-health applications, may offer increasing opportunities for the collection and analysis of real-world data in order to achieve a more automatic, objective, and granular assessment with respect to the current clinical practice.
However, existing solutions exploit AI mainly for image-based food recognition and food volume assessment in order to estimate nutrient and calories intake \citep{liu2017new}, while they do not integrate appropriate data analysis and inference pipelines to investigate the impact of heterogeneous data on the individual nutritional status over the time.
Machine Learning (ML) and Deep Learning (DL) algorithms may provide effective and scalable solutions for care providers support in the early identification and management of subjects at risk of malnutrition \citep{sharma2020malnutrition}.
\\ 
However, the best performing models are often too complex to be self-explaining, and this lack of transparency and interpretability generally prevents from the acceptance and spread in the clinical practice.
To this aim, Explainable AI (XAI) comes into play to improve human understanding and confidence in AI-empowered predictions and decisions \citep{arrieta2020explainable}, thus representing a natural step towards {\it Trustworthy AI}.
XAI approaches generally include developing novel inherently interpretable AI algorithms or complementing complex, {\it\enquote{black-box}} models with \textit{post-hoc} explanations.
However, the application of XAI methodologies is not sufficient as standalone solution to support informed and confident decisions if an extensive assessment of the generated explanations is neglected to close the loop, especially in healthcare applications.
Clinical validation, consistency assessment, as well as quality assessment currently represent the key features to ensure consumable explanations for both patients and clinicians \citep{di2022explainable}.
Clinical validation is an essential requirement in healthcare to add credibility to the final model by comparing data-driven versus evidence-based knowledge, while evaluating the level of agreement among the explanations provided by multiple methods, may provide some preliminary understanding of the system's stability and robustness. Eventually, both objective and human-centered quality evaluations should be integrated to determine which form of offered explanations is best suited for the target audience \citep{doshi2017towards}.

\subsection{Contribution}
In this paper, we present a novel framework for explainable and early malnutrition risk prediction, with particular reference to institutionalised older adults.
The framework is built on the top of a m-health application specifically designed and deployed to monitor older adults in Long-Term Care (LTC) settings in order to enable a semi-continuous malnutrition risk screening (i.e., on a weekly basis), with the ultimate goal of aiding care providers in early subject identification and management.
This work represents an extension of our previous pilot study published in \cite{di2021malnutrition}.
The solution proposed in this paper provides several additions and improvements, including a larger data collection, the integration of clinical information and heterogeneous monitoring data, additional ML algorithms, as well as the application and assessment of several benchmark XAI methods to explain the best performing yet complex models.
Model development and evaluation aims to achieve accurate classification while ensuring result generalisability, thanks to larger training sets and to a comprehensive performance analysis that encompasses both subject-independent and personalised predictions.
Then, we extracted explanations to enhance interpretability of the best performing models at global stage, followed by a two-step explanation assessment stage. Specifically, we first performed an objective assessment of the level of agreement between explanations generated through different methods for each model separately. Then, we performed a preliminary clinical validation to verify that the input-target relationships learned for the most relevant predictors are in line with the current evidence-based assessment.

\section{Background}\label{sec:background}
Most of the existing AI solutions for nutritional assessment focus on automatic food recognition and/or food volume reconstruction from smartphone images or camera sensors, as this enables to combine the estimated consumption of each detected food item with reference food composition databases to evaluate daily nutrient and caloric intakes.
These systems generally require some level of interactivity and understanding by end users to ensure a good usability, and hence they are mainly thought for supporting healthy subjects in independent living scenarios.
For instance, \cite{elfert2021deep} proposed a system to support older people living at home to keep a digital nutrition diary. The framework exploits a SSD-MobileNet V$2$ Convolutional Neural Network (CNN) architecture to classify smartphone food pictures according to specific food macro-categories, then it implements an interactive interview mode to allow end user to select the correct food item within the detected group. Finally, food intake assessment is based on standard household quantities entered manually by the user.
On the other hand, few solutions have been specifically designed for clinical contexts, such as hospitals or LTC institutions (with particular reference to older adults), since they must provide a fully automated, reliable, time- and cost-efficient monitoring that can be easily integrated within the daily routine of clinical/nursing care personnel \citep{doulah2019systematic}.
\cite{lu2020artificial} proposed a sequential approach to detect food intake in hospitalised patients by processing $3$-D RGB images of the food tray, captured before and after meal consumption through depth camera sensors.
The system includes a Multi-Task Contextual Network for food segmentation, a few-shot learning-based classifier for food recognition, and finally an algorithm for $3$-D surface construction.
Similarly, \cite{ruenin2020system} proposed an automatic system for calories intake estimate in hospitalised older subjects based on a two-step CNN approach for food classification and food weight assessment. More recently, \cite{pfisterer2022automated} proposed a novel deep convolutional encoder-decoder network with depth refinement using RGB-D camera images to estimate the remaining volume of mixed food items contained in LTC dishes.
\\
This type of systems do not generally investigate the correlation between nutritional intake and nutritional status over the time, especially in terms of malnutrition. Moreover, they do not include other heterogeneous sensing data (i.e., behavioural, anthropometric, physiological etc...) that may represent potential risk factors. Currently, the majority of existing AI frameworks for predicting nutritional disorders are mainly based on static demographic/clinical information collected through national health surveys, as well as on biochemical markers collected during few examinations. As a result, the potential integration with m-health data to enable an accurate and more continuous assessment is still largely unexplored.
In addition, specific solutions for older adults are really limited.
\cite{kang2019sarcopenia} proposed a knowledge-based feature selection and a Random Forest (RF) classifier to predict sarcopenia (i.e., muscle mass loss) in older and hospitalised patients, using this outcome as proxy of malnutrition condition. Here, input data mainly consist of blood test parameters, whereas nutrient intake information are retrieved by a national, cross-sectional dietary survey. \cite{panagoulias2021nutritional} considered body weight as one of the most important indicators of nutritional health, proposing an Artificial Neural Network (ANN) model for predicting Body Mass Index (BMI) category (i.e., underweight, normal, and overweight) from metabolomics data (i.e., metabolite concentrations in plasma) in a sample of $6413$ subjects belonging to different ethnicity and age groups.
\cite{yin2021fusion} developed a clinical decision support system to detect and grade malnutrition in cancer patients using $17$ core features including demographic, clinical, anthropometric, and lab test data. The system implements a multi-stage K-means clustering to automatically detect different malnutrition severity levels within observed data, then multiple ML models have been tested for malnutrition prediction.
Eventually, \cite{larburu2022key} proposed a AI framework to characterise malnutrition risk in hospitalised older women using a set of demographic, clinical, and laboratory information, as well as hospitalisation factors. They first developed ML models to distinguish between no risk and risk conditions, then additional models have been developed to detect low versus high risk in subjects previously classified at risk. For both tasks, RF and gradient boosting models achieved the best results.
\\ 
As far as explainability is concerned, very few works exploit XAI to understand malnutrition and nutrition-related disorder predictions. Many applications often rely on knowledge-based approaches or on simpler and intrinsically interpretable ML models (e.g., linear models), often imposing a trade-off between performance and interpretability.
For instance, \cite{cioara2018expert} developed an ontology-based expert system for early malnutrition detection in older adults, whereas \cite{azevedo2022diagnostic} developed sex-specific multivariate logistic regression models to evaluate the relevance of several clinical, laboratory, and body composition variables in predicting sarcopenic obesity in older adults. However, few solutions also exist for explaining well performing yet \textit{\enquote{black-box}} models.
\cite{benmohammed2022metabolic} proposed a novel AI-based approach for metabolic syndrome (MetS) screening in adolescents from anthropometric measurements and blood tests, using XAI to estimate the learned decision function. Specifically, RF-based Mean Decrease in Impurity (MDI) method has been used in combination with medical expertise for feature selection, then only $4$ features (i.e., age, waist circumference, mean blood pressure, and triglycerides) have been considered to determine a polynomial formula of the decision function for the best performing ANN architecture by using soft margin technique. Classification performance obtained by using the learned function outperformed $5$ reference clinical definitions used for Mets diagnosis.
\cite{pang2019understanding} proposed a XAI approach to analyse the main risk factors associated to early childhood obesity. To this aim, they trained an eXtreme Gradient Boosting (XGBoost) model with Electronic Health Record (EHR) data collected by $>860$ pediatric subjects, then they applied SHapley Additive ExPlanations (SHAP) technique \citep{lundberg2017unified} to get global model explanations.
The obtained results show that well-established obesity risk factors such as weight, height, weight-for-height, geographic location, race and ethnicity are detected among the most important features. On the other hand, SHAP also highlighted additional relevant features related to human metabolism, such as body temperature and respiration rate, which might represent emergent biomarkers requiring further investigation in future research. Furthermore, \cite{shi2022explainable} developed several ML models to predict the occurrence of post-operative malnutrition in children with congenital heart disease within $1$ year after surgery. Specifically, the primary outcome of interest was the occurrence of underweight status, while the secondary outcomes were stunted and wasting
status, respectively. To this aim, several ML models have been trained to predict the above outcomes separately by using information collected by intra-operative and follow-up EHR, and XGBoost achieved the best results in all cases. Finally, feature permutation methods and SHAP have been used to determine the most impactful features for each malnutrition-related condition.
\\
Eventually, to the best of our knowledge, there is no study available in the literature that conducts an extensive assessment of relevance and quality of explanations related to AI-based malnutrition prediction from both clinical and m-health data (e.g., nutritional and body composition data), and especially in terms of clinical soundness.

\section{Material and Methods}
This section describes the full AI pipeline, starting from data collection and preparation, up to ML model selection and evaluation. 

\subsection{Population and data collection}
In \cite{di2021malnutrition}, we collected data from different groups of older adults housed in an Italian LTC facility\footnote{http://www.icareviareggio.it/} within $3$ separate trial periods (T$1$-T$3$), ranging from March $2018$ to August $2020$ ($52$ weeks in total).
To this aim, we designed and deployed a simple yet efficient m-health application, named \textit{DoEatWell} (DEW), to enable a long-term monitoring of this particular population category in clinical settings.
A complete description of DEW has been already reported in \cite{delmastro2018long}. To summarise, here we briefly present DEW main functionalities:
\begin{itemize}
    \item Food tracking: DEW enables tracking users' preferences for the main daily meals;
    \item Food intake assessment: DEW implements a qualitative visual food intake assessment through a simple and quick survey to be filled in at the end of each meal (LTC guests can answer it autonomously or be supported by a care giver);
    \item Body composition assessment: DEW is integrated with a smart bioimpedance scale through Bluetooth Low Energy interfaces to collect weight and body composition statistics.
\end{itemize}

For the current study, we extended our dataset by collecting data from two additional groups of $23$ subjects in two different trial periods in $2021$, namely T$4$ and T$5$, for a total of $44$ weeks (i.e, $10$ months).
More details about the number of enrolled subjects, number of monitoring weeks, and number of collected observations for each trial period are reported in Table \ref{tab_trial_periods}.
As it may be noticed,  both group size and monitoring duration differ across the trial periods, ranging from $9$ to $23$ subjects and from $13$ to $23$ weeks, respectively. 
Trial periods are spaced out over time, ranging from a few months (e.g., T$4$-T$5$) to up to $3$ years (e.g., T$1$-T$4$), with a limited overlap between samples in consecutive periods.  
The degree of overlap, presented in Table \ref{tab_trial_periods}, is indicated as the ratio between the number of subjects who participated in both the current (T$_i$) and previous (T$_{i-1}$) monitoring periods and the group size in the current period. 
This turn-over may be partially attributed to the elapsed time between the different trial periods, but it is also a typical condition in LTC settings as frequent subject's drop-outs occur due to sudden and unpredictable events (e.g., health aggravations, transfer to other LTC institutions/hospitals), often preventing long-term monitoring of the same subjects. 
As a result, each trial period can be considered independent, with longitudinal (weekly) observational data collected only for subjects participating in the respective trial. 
Furthermore, only a subset of subjects was typically able to perform regular body composition assessments in each trial period using the smart scale because of temporary or permanent physical impairments. Therefore, we analysed subjects with and without body composition data separately.
\\
Table \ref{tab_subject_data} provides an overview of the collected datasets for each subject, showing the number of weekly observations gathered in each trial period as well as the total count. Notably, most subjects participated in one to three trial periods, with only a few subjects (i.e., subject $\#7$, $\#11$, $\#13$, and $\#14$) achieving a complete monitoring. As a result, the individual dataset size exhibits a significant variability, ranging from $4$ to $79$ observations, and the data sequences are often irregularly sampled across distant time intervals.
Eventually, the overall datasets obtained by merging all monitoring periods consist of approximately $2$ years (i.e., $96$ weeks) of data collected from a total of $42$ and $27$ different subjects, respectively. 
This data were used to develop and evaluate several benchmark ML models to detect the early onset (i.e., on a weekly basis) of malnutrition risk condition compared to a healthy nutritional status, leveraging the clinical assessment obtained at the end of each month through MNA-SF for dataset labeling.

\begin{table}[!htbp]
\caption{Overview of data collected in each trial period and of the resulting datasets.}
\label{tab_trial_periods}%
\begin{minipage}{\textwidth}
\begin{threeparttable}
\centering
\resizebox{\textwidth}{!}{
\begin{tabular}{lcccccccc}
\toprule
\textbf{Trial Period}&\textbf{Body compo (Y/N)}&\textbf{\# weeks}&\textbf{\# subjects}&\textbf{sample overlap}&\textbf{\# collected obs}&\textbf{invalid nutritional data ($\%$)}&\textbf{missing body compo data ($\%$)}&\textbf{Class \% (Normal/Risk) }\\
\midrule
T$1$ (Mar-Sep $2018$)& N& $22$& $15$& N.A.& $306$& $10.5$& N.A.& $82.3$/$17.7$\\ 
\midrule
T$1$& Y& $22$& $8$& N.A.& $160$& $6.2$& $28.7$& $83.1$/$16.9$\\
\midrule
T$2$ (Dec $2019$- Mar $2020$)\tnote{\textdagger} & N& $17$& $9$& $5/9$ & $144$&\textcolor{red}{\textbf{85.4}}& N.A.& $64.6$/$35.4$\\
\midrule
T$2$\tnote{\textdagger} &Y& $17$& $9$& $4/9$ & $144$& \textcolor{red}{\textbf{85.4}}& $22.9$& $64.6$/$35.4$\\
\midrule
T$3$ (Jun-Aug $2020$)& N& $13$& $23$& $8/23$& $286$& $7.0$& N.A.& $60.8$/$29.2$\\
\midrule
T$3$ & Y& $13$& $12$& $8/12$& $156$& $7.0$& $16.7$& $75.0$/$25.0$\\
\midrule
T$4$ (Jan-Apr $2021$)& N& $17$& $23$& $18/23$& $391$& $2.6$& N.A.& $57.5$/$42.5$\\
\midrule
T$4$& Y& $17$& $15$& $10/15$& $255$& $0.0$& $16.9$& $61.2$/$38.8$\\
\midrule
T$5$ (Jul-Dec $2021$)& N& $27$& $23$& $16/23$& $616$& $8.9$& N.A.&$68.8$/$31.2$\\
\midrule
T$5$& Y& $27$& $18$& $13/18$& $486$& $7.2$& $17.5$&$65.8$/$34.2$\\
\midrule
All (T$1$-T$5$)& N& $96$& $42$& N.A.& $1599$& $7.3$\tnote{*}& N.A& $67.2$/$32.8$\tnote{*}\\
\midrule
All (T$1$-T$5$)& Y& $96$ & $27$& N.A.& $1057$& $5.3$\tnote{*}& $22.0$\tnote{*}& $68.7$/$31.3$\tnote{*}\\
\bottomrule
\end{tabular}}
\begin{tablenotes}\footnotesize
\item[\textdagger] \textit{T$2$ observations have been discarded due to the high \% of invalid nutritional data.}
\item[*] \textit{Percentages are computed without including T$2$ data.}
\end{tablenotes}
\end{threeparttable}
\end{minipage}
\end{table}

\begin{table}[!htbp]
\begin{minipage}{\textwidth}
\centering
\caption{Subject-specific datasets without (left) and with (right) body composition assessment.}
\label{tab_subject_data}%
\resizebox{\textwidth}{!}{
\begin{tabular}{cc}
    \begin{tabular}[t]{ccccccc}
    \toprule
        \textbf{Subject ID} & \textbf{T1} & \textbf{T3} & \textbf{T4} & \textbf{T5} & \textbf{Total (T1-T5)} & \textbf{Class \% (Normal/Risk) }\\ 
        \midrule
        1 & 22 & 13 & ~ & ~ & 35 & 100/0\\ 
        \midrule
        2 & 22 & 13 & 17 & ~ & 52& \textbf{67.3/32.7} \\
        \midrule
        3 & 22 & 13 & ~ & ~ & 35& \textbf{88.5/ 11.5} \\ 
        \midrule
        \textcolor{red}{\textbf{4}} & 22 & ~ & ~ & ~ & \textcolor{red}{\textbf{22}}& 100/0 \\ 
        \midrule
        \textcolor{red}{\textbf{5}} & 22 & ~ & ~ & ~ & \textcolor{red}{\textbf{22}}& 0/100 \\ 
        \midrule
        \textcolor{red}{\textbf{6}} & 22 & ~ & ~ & ~ &\textcolor{red}{\textbf{22}} & 100/0\\ 
        \midrule
        7 & 22 & 13 & 17 & 22 & 74& \textbf{89.2/10.8} \\
        \midrule
       \textcolor{red}{\textbf{8}} & 18 & ~ & ~ & ~ & \textcolor{red}{\textbf{18}} & 100/0 \\ 
        \midrule
        \textcolor{red}{\textbf{9}} & 10 & ~ & ~ & ~ & \textcolor{red}{\textbf{10}}& 100/0 \\ 
        \midrule
        10 & 22 & ~ & ~ & ~ & \textcolor{red}{\textbf{22}} & 77.2/22.8\\ 
        \midrule
        11 & 22 & 13 & 17 & 27 & 79& \textbf{26.6/ 73.4}\\ 
        \midrule
        \textcolor{red}{\textbf{12}} & 14 & ~ & ~ & ~ & \textcolor{red}{\textbf{14}}& 100/0\\ 
        \midrule
        13 & 22 & 13 & 17 & 27 & 79& 100/0 \\ 
        \midrule
        14 & 22 & 13 & 17 & 27 & 79& 100/0 \\ 
        \midrule
        \textcolor{red}{\textbf{15}} & 22 & ~ & ~ & ~ & \textcolor{red}{\textbf{22}}& 0/100 \\ 
        \midrule
        16 & ~ & 13 & 17 & ~ & 30& \textbf{71/29} \\ 
        \midrule
        17 & ~ & 13 & 17 & 27 &57& 0/100 \\ 
        \midrule
        18 & ~ & 13 & 17 & 27 & 57 & 100/0\\ 
        \midrule
        19 & ~ & 13 & 17 & 27 & 57 & 100/0\\ 
        \midrule
        20 & ~& 13 & 17 & ~ & 30& 0/100 \\ 
        \midrule
        21 & ~ & 13 & 17 & 27 & 57& \textbf{93/7} \\ 
        \midrule
        \textcolor{red}{\textbf{22}} & ~ & 9 & ~ & ~ & \textcolor{red}{\textbf{9}} &100/0 \\ 
        \midrule
        23 & ~ & 13 & 17 & 27 & 57& 0/100 \\
        \midrule
        24 & ~ & 13 & 17 & 27 & 57& \textbf{15.8/ 84.2} \\ 
        \midrule
        \textcolor{red}{\textbf{25}} & ~ & 4 & ~ & ~ & \textcolor{red}{\textbf{4}}& 100/0 \\ 
        \midrule
        26 & ~ & 13 & 17 & 27 &57& 100/0 \\ 
        \midrule
        27 & ~ & 13 & 17 & 27 & 57 & \textbf{70.2/19.8}\\ 
        \midrule
        28 & ~ & 13 & 17 & ~ & 30& 0/100 \\
        \midrule
        29 & ~ & 13 & ~ & 27 & 40 & \textbf{45/55} \\ 
        \midrule
        30 & ~ & 13 & 17 & 27 & 57& 100/0 \\ \midrule
        31 & ~& 13 & 17 & ~ & 30& \textbf{30/70} \\ 
        \midrule
        \textcolor{red}{\textbf{32}} & ~& ~ & 17 & ~ & \textcolor{red}{\textbf{17}}& 0/100 \\ 
        \midrule
        33 & ~ & ~ & 17 & 27 & 44 & 100/0 \\ 
        \midrule
        34 & ~ & ~ & 17 & 27 & 44 & 100/0 \\ 
        \midrule
        35 & ~ & ~ & 17 & 27 & 44 & \textbf{68.2/ 31.8} \\ 
        \midrule
        \textcolor{red}{\textbf{36}} & ~ & ~ & 17 & ~ & \textcolor{red}{\textbf{17}}& 53/47 \\ 
        \midrule
        37 & ~ & ~ & ~ & 27 & 27& 0/100 \\ 
        \midrule
        38 & ~ & ~ & ~ & 27 & 27& 100/0 \\ 
        \midrule
        39 & ~ & ~ & ~ & 27 & 27 & 100/0 \\ 
        \midrule
        40 & ~ & ~ & ~ & 27 & 27& 100/0 \\ 
        \midrule
        41 & ~ & ~ & ~ & 27 & 27& 0/100 \\ 
        \midrule
        42 & ~ & ~ & ~ & 27 & 27& 100/0 \\ 
        \bottomrule
    \end{tabular}&
    \begin{tabular}[t]{ccccccc}
    \toprule
        \textbf{Subject ID} & \textbf{T1} & \textbf{T3}& \textbf{T4}& \textbf{T5}& \textbf{Total (T1-T5)}& \textbf{Class \% (Normal/Risk) } \\ 
        \midrule
        1 & 22 & 13 & ~ & ~ & 35& 100/0 \\ 
        \midrule
        3 & 22 & 13 & ~ & ~ & 35& \textbf{88.5/11.5} \\ 
        \midrule
        \textcolor{red}{\textbf{4}}& 22 & ~ & ~ & ~ & \textcolor{red}{\textbf{22}}& 100/0 \\ 
        \midrule
        \textcolor{red}{\textbf{5}} & 22 & ~ & ~ & ~ &\textcolor{red}{\textbf{22}}& 0/100 \\
        \midrule
        \textcolor{red}{\textbf{8}} & 18 & ~ & ~ & ~ &\textcolor{red}{\textbf{18}}& 100/0 \\ 
        \midrule
        \textcolor{red}{\textbf{9}} & 10 & ~ & ~ & ~ &\textcolor{red}{\textbf{10}}& 100/0 \\ 
        \midrule
        \textcolor{red}{\textbf{10}} & 22 & ~ & ~ & ~ & \textcolor{red}{\textbf{22}}& 77.2/22.8 \\ 
        \midrule
        14 & 22 & 13 & 17 & 27 & 79 & 100/0\\ 
        \midrule
        11 & ~ & 13 & 17 & 27 & 57 & \textbf{26.6/73.4} \\ 
        \midrule
        16 & ~ & 13 & 17 & ~ & 30 & \textbf{71/29} \\ 
        \midrule
        17 & ~ & 13 & 17 & 27 & 57 & 0/100\\ 
        \midrule
        18 & ~ & 13 & 17 & 27 & 57 & 100/0 \\ 
        \midrule
        19 & ~ & 13 & 17 & 27 &57 & 100/0\\ 
        \midrule
        21 & ~ & 13 & 17 & 27 & 57 & \textbf{93/7} \\ 
        \midrule
        23 & ~ & 13 & 17 & 27 & 57 & 0/100\\ 
        \midrule
        26 & ~ & 13 & 17 & 27 & 57& 100/0\\ 
        \midrule
        30 & ~ & 13 & 17 & 27 & 57 & 100/0\\ 
        \midrule
        32 & ~ & ~ & 17 & ~ & \textcolor{red}{\textbf{17}}& 0/100 \\ 
        \midrule
        24 & ~ & ~ & 17 & 27 & 44 & \textbf{15.8/84.2} \\ 
        \midrule
        33 & ~ & ~ & 17 & 27 & 44 & 100/0 \\ 
        \midrule
        34 & ~ & ~ & 17 & 27 & 44 & 100/0 \\ 
        \midrule
        35 & ~ & ~ & 17 & 27 & 44 & \textbf{68.2/31.8} \\ 
        \midrule
        37 & ~ & ~ & ~ & 27 & 27 & 0/100 \\ 
        \midrule
        39 & ~& ~ & ~ & 27 & 27 & 100/0 \\ 
        \midrule
        40 & ~ & ~ & ~ & 27 &27 & 100/0\\ 
        \midrule
        41 & ~ & ~ & ~ & 27 & 27 & 0/100 \\ 
        \midrule
        42 & ~ & ~ & ~ & 27 & 27 & 100/0 \\ 
        \bottomrule
    \end{tabular}\tabularnewline
    \end{tabular}}
    \end{minipage}
\end{table}

\subsection{Feature re-engineering}\label{sec:feat_reing}
In collaboration with a medical specialist, we revised the set of input features with the ultimate goal of boosting model predictive performances, as well as analysing the impact of specific malnutrition indicators.
Table \ref{tab_features} shows the complete list of eligible input predictors, indicating whether each of them has been used in our previous (i.e., T$1$-T$3$) and/or current (i.e., T$1$-T$5$) implementations.
\\
As far as body composition is concerned, we initially considered only the Fat Mass Index (FMI), a measure of relative fat mass content whose reduction is known to be associated to an increased malnutrition risk in the older population \citep{tomlinson2019body}.
However, malnutrition is also often accompanied by muscle mass loss in older adults, especially in case of protein-energy deficiencies \citep{sieber2019malnutrition}, even with fat mass values within or above normality range in some cases. Therefore, quantitative muscle mass assessment is important during ageing as well, and bioelectrical impedance analysis (BIA) represents the most feasible method in routine clinical practice with respect to dual-energy x-ray absorptiometry and computerized tomography \citep{barazzoni2022guidance}. Unfortunately, the smart biompendance scale used in our study just provides an estimate of body lean mass (also known as fat-free mass), which basically includes all non-fat components along with muscle mass in the total computation, such as bones, water, skin, and internal organs. Therefore, the actual relationship between muscle mass variation and malnutrition risk may not be discerned accurately. For this reason, we selected BMI (i.e., body weight with respect to height) as a more general indicator to investigate the impact of the overall weight variation due to fat and/or muscle mass reduction on malnutrition.
In addition, we included Basal Metabolic Rate (BMR, expressed in KCal/day) to analyse the metabolic response to a chronic or prolonged inadequate diet. Indirect calorimetry is the reference standard method by which resting metabolism is derived from measurements of oxygen consumption and carbon dioxide production \citep{ferrannini1988theoretical}; however, smart scale devices can automatically estimate BMR upon body weight measurement in a quicker and less invasive way, although less accurate, by using Harris-Benedict equations and successive refinements \citep{mifflin1990new}.
Finally, we also selected total body water percentage to assess the impact of body hydration on the malnutrition condition.
\\
For each subject, the above body composition statistics have been averaged across all the available measurements in each monitoring week. Moreover, we introduced patient clinical attributes that may represent relevant auxiliary inputs. This information is stored in DEW user health profile, and it can be updated by the medical/nursing personnel directly through the application.
First, we added sex and age of participants as they are among the most important confounding variables in clinical/epidemiological studies \citep{ward2021explainable}.
We also added a Boolean variable to indicate whether the subject performed regular physical training and/or rehabilitation activities, defined by physiotherapists, during the weekly LTC routine. Then, we considered the number of comorbidities (i.e. coexisting chronic disorders) and the number of therapies (i.e., pharmacological treatments). Specifically, we selected only chronic disorders with a potential impact on the individual nutritional status, such as diabetes, hypertension and cardiovascular diseases, liver disease, gastrointestinal diseases, and many others. This information did not generally change for each subject in each trial period unless the onset of new disease and/or the supply of additional drugs.
Moreover, clinical malnutrition assessment also includes an evaluation of cognitive impairments and dementia (and also depression), as they negatively impact on dietary habits. For this reason, we introduced the outcomes of the Mini Mental State Examination (MMSE) \citep{tombaugh1992mini} in order to correlate the individual cognitive status with the risk of developing malnutrition.
MMSE provides a score ranging from $0$ to $30$, with scores $\ge24$ indicating a normal cognitive status, whereas scores within the range [$18$-$23.5$] indicate Mild Cognitive Impairment (MCI). Finally, scores below $18$ indicate a severe impairment of cognitive abilities, often leading to dementia.
MMSE is generally administered by a medical specialist periodically, with a minimum period of $3$ months. Therefore, for each subject we considered the most recent examination for each trial period, updating scores as soon as new evaluations are available.
\\
Regarding dietary assessment, we mainly focus on estimating the intake of four main macro-nutrients, namely cereals, animal proteins, vegetables, and fruit. This choice is motivated by the fact that the daily menus offered by the LTC facility internal canteen are designed to follow the guidelines of the Mediterranean diet, therefore these macro-nutrients are considered the necessary components of the \enquote{Healthy Eating Plate} \cite{willett2017eat} for a balanced diet.
However, we revised the daily intake calculation in order to get more accurate quantitative estimates.
In our previous work, the daily intake of each macro-nutrient was expressed as percentage with respect to a recommended daily allowance, using weights defined a priori by a medical specialist to account for the importance of each macro-nutrient in each course (i.e., first course, second course, side dish, fruit/dessert) for lunch and dinner meals. Please refer to \cite{di2021malnutrition} for the complete algorithm formulation.
Here, we introduced the nominal portion (expressed in grams) of each food item included in the menu. This information has been added to the nutritional fact database stored in DEW cloud module. Then, we combined the received portion of each food item with the corresponding consumption and the corresponding nutritional composition to estimate the daily intake of each macro-nutrient. Daily intake values have been finally averaged across all the valid monitoring days (i.e., the days that present both lunch and dinner surveys filled in) in each week.
On the other hand, we discarded the variability indexes related to macro-nutrient intakes (i.e., the mean of the successive differences between daily values), as average daily intake estimates may represent a sufficient information to characterise the users' dietary habits on a weekly basis.
\\
Finally, we also considered behavioral indexes related to meal completeness and variability that may potentially track unhealthy dietary habits.
For instance, subject may have difficulties in consuming all the food items included in the proposed daily menus due to hyporexia (i.e., decreased appetite) or anorexia, resulting in a reduced nutrient and energy intake. Moreover, older adults often exhibit a lack of response to diet monotony \citep{pelchat2000dietary} due to several problems such as apathy, depression, smell and taste loss, with a significant impact on the intake frequency of the main food nutrients. However, these behavioral markers may not be well suited for our specific reference scenario, in which subjects with limited cognitive abilities are generally guided by the nursing care personnel for food selection.
Such external interventions may provide some corrections to the user behaviour, therefore limiting the ability of the above behavioural markers to faithfully measure unhealthy dietary habits associated with malnutrition risk over the time.
Moreover, ML models trained by only combining nutritional and behavioural features (without the support of body composition assessment) reached unsatisfactory predictive performances.
As a result, the behavioural features listed in Table \ref{tab_features} have been removed in the current study, leaving their possible application to future studies in less controlled contexts, such as independent living scenarios. 

\begin{table}[!htbp]
\begin{minipage}{\textwidth}
\centering
\begin{threeparttable}
\caption{List of input features.}\label{tab_features}%
\begin{tabular}{lllll}
\toprule
\textbf{Feature Name} & \textbf{Feature Type}&  \textbf{Data type}&\textbf{T1-T3}&\textbf{T1-T5}\\
\midrule
Fat Mass Index (FMI) & Body composition &Numerical&\checkmark&\checkmark\\
\midrule
Body Mass Index (BMI) & Body composition &Numerical&$\times$&\checkmark\\
\midrule
Basal Metabolic Rate (BMR) & Body composition &Numerical&$\times$&\checkmark\\
\midrule
Body Water Percentage & Body composition &Numerical&$\times$&\checkmark\\
\midrule
Age&Clinical& Numerical&$\times$&\checkmark\\
\midrule
Sex& Clinical& Categorical&$\times$&\checkmark\\
\midrule
Physical Activity&Clinical& Boolean&$\times$&\checkmark\\
\midrule
Mini Mental State Examination (MMSE)&Clinical& Numerical&$\times$&\checkmark\\
\midrule
Comorbidities &Clinical&Numerical&$\times$&\checkmark\\
\midrule
Therapy&Clinical&Numerical&$\times$&\checkmark\\
\midrule
Avg. Daily Cereal Intake\tnote{$\ast$}& Nutritional&Numerical&\checkmark&\checkmark\\
\midrule
Avg. Daily Protein Intake\tnote{$\ast$}& Nutritional&Numerical&\checkmark&\checkmark\\
\midrule
Avg. Daily Vegetable Intake\tnote{$\ast$}& Nutritional&Numerical&\checkmark&\checkmark\\
\midrule
Avg. Daily Fruit Intake\tnote{$\ast$}& Nutritional&Numerical&\checkmark&\checkmark\\
\midrule
Avg. Daily Cereal Intake Variability& Nutritional&Numerical&\checkmark&$\times$\\
\midrule
Avg. Daily Protein Intake Variability& Nutritional&Numerical&\checkmark&$\times$\\
\midrule
Avg. Daily Vegetable Intake Variability& Nutritional&Numerical&\checkmark&$\times$\\
\midrule
Avg. Daily Fruit Intake Variability& Nutritional&Numerical&\checkmark&$\times$\\
\midrule
Lunch Completeness& Behavioural&Numerical&\checkmark&$\times$\\
\midrule
Dinner Completeness& Behavioural&Numerical&\checkmark&$\times$\\
\midrule
Lunch Variability& Behavioural&Numerical&\checkmark&$\times$\\
\midrule
Dinner Variability& Behavioural&Numerical&\checkmark&$\times$\\
\midrule
First Course Variability& Behavioural&Numerical&\checkmark&$\times$\\
\midrule
Second Course Variability& Behavioural&Numerical&\checkmark&$\times$\\
\midrule
Side Dish Variability& Behavioural&Numerical&\checkmark&$\times$\\
\bottomrule
\end{tabular}
\begin{tablenotes}\footnotesize
\item [$\ast$] The algorithm for daily macro-nutrient intake calculation has been revised for T$1$-T$5$ models.\end{tablenotes}
\end{threeparttable}
\end{minipage}
\end{table}

\subsection{Model development and evaluation}
We first analysed data quality in order to detect and replace missing and/or unreliable data.
For body composition assessment, if there was no available measurements within a given week, we replaced the corresponding observation with the average of the two closest measurements over time.
The same strategy has been applied for nutritional data, increasing the threshold of valid monitoring days in a week with respect to our previous study from $3$ to $5$ in order to get more accurate and reliable average daily nutrient intake estimates. However, as highlighted in Table \ref{tab_trial_periods}, this choice led to a high percentage of invalid observations in T$2$ ($85.4\%$), due to poor service adherence. As a result, we did not apply data imputation and discarded all data collected in this trial period, yet reducing the overall number of analysed observations with respect to those theoretically available.
\\
The resulting dataset with only clinical and nutritional variables consists of $1599$ observations and $10$ predictors, whereas the dataset including body composition assessment has $1057$ samples and $14$ input features. In both cases, the size of the new datasets has grown by a factor $>2x$ compared to those collected in our previous work.
The new datasets also exhibit a very similar class imbalance (approximately $2$:$1$), with malnutrition risk as the under-represented condition, ranging from $31.3\%$ (with body composition assessment) to $32.8\%$ (without body composition assessment) of the total observations.
To account for dataset imbalance, we compared model training in unbalanced settings with respect to Synthetic Minority Oversampling Technique (SMOTE) \citep{chawla2002smote}, as well as against cost-sensitive learning \citep{elkan2001foundations} approaches penalising false positive and false negative misclassifications with a different degree.
Moreover, we also evaluated Balanced Random Forest (BRF) and Random Undersampling Boosting (RUSBoost) as ad-hoc algorithms to naturally handle unbalanced data. For both models, each weak learner in the ensemble is trained using a balanced subset randomly drawn from the original training data.
We applied specific preprocessing to input predictors according to the data type. Specifically, we converted the Boolean variable (i.e. Physical Activity) into $0$-$1$ values, then we applied One-Hot Encoding to the categorical variable (i.e., Sex). Finally, we applied standardisation to all numerical predictors. 
\\ 
We conducted a two-stage model validation process in order to enhance the reliability of our findings regarding result generalisability, as well as to better support the identification of the best performing models.
To this aim, we first evaluated subject-independent model predictions by dividing the datasets into random, stratified hold-out partitions ($70\%$ training, $30\%$ test). This process was repeated $10$ times to ensure robustness in the evaluation. The purpose of this stage was to obtain a comprehensive understanding of how the models perform across different data splits.
Then, we employed Leave-One-Subject-Out Cross-Validation (LOSO-CV) to evaluate personalised predictions. This approach involves systematically leaving out one subject at a time from the training set and evaluating the model performances on the left-out subject. LOSO-CV allows for a more fine-grained assessment of how well the models generalise to individual subjects, considering their unique characteristics and patterns.
For both validation stages, we trained models using $10$-fold CV with Bayesian optimisation \citep{snoek2012practical} for hyperparameter tuning ($M$=$60$ optimisation rounds), using F$1$-score as scoring metric for model selection in case of unbalanced training, accuracy otherwise.
Moreover, we replaced AdaBoost (AB) algorithm with two different gradient boosting implementations, namely XGBoost and Light Gradient Boosting Machine (LightGBM). These boosting algorithms exploit gradient descent method to continuously minimise the log-loss function (i.e., binary cross-entropy) in order to find the optimal point, thus potentially leading to better outcomes with respect to AB. In addition, they introduce regularisation parameters to reduce overfitting, and they are both optimised for speed with respect to AB (with LightGBM faster than XGBoost).
\\ 
For what concerns subject-independent predictions, we evaluated aggregated performances over multiple runs, including accuracy, F$1$-score, and Area Under the ROC Curve (AUC) to provide a comprehensive and fair analysis that accounts for dataset imbalance.
For each model, we compared these metrics with respect to the different data imbalance management methods to detect significant differences. To this aim, we first verified variance homogeneity through Bartlett's test, then we applied $1$-way Analysis of Variance (ANOVA) followed by Tukey's Honestly Significant Difference (HSD) test for post-hoc comparisons, when necessary.
In addition, we selected the top-performing configuration for each model type as the reference for comparison against the corresponding best implementations obtained in our preliminary study to check for performance gains.
\\
We applied the same statistical analysis also to LOSO-CV results, by aggregating model performances across all subjects. However, as shown in Table \ref{tab_subject_data}, each monitored subject provides a varying number of observations, ranging from $4$ to $79$. Consequently, we set a minimum threshold of $26$ weeks to exclude subjects with limited monitoring (i.e., approximately less than $6$ months). As a result, $12$ out of $42$ subjects without body composition data and $6$ out of $27$ subjects with body composition data were excluded. The ID and number of observations for the removed subjects are highlighted in red in the corresponding tables. Moreover, during the monitoring period(s), only a few selected subjects exhibited variations in their nutritional status. Specifically, $11$ out of $30$ subjects ($36.7\%$) with body composition data and $6$ out of $21$ subjects ($28.5\%$) without body composition data experienced changes in their nutritional status. The overall class distribution for these subjects is indicated in bold in the corresponding tables. Conversely, the majority of subjects had observations belonging to a single class (either \textit{Normal} or \textit{Malnutrition Risk}).
In such cases, precision, recall, and F$1$-score are ill-defined, and AUC cannot be computed. Therefore, we focused on classification accuracy when evaluating the results of LOSO-CV.
\\
Subsequently, we compared accuracy values obtained from subject-independent and personalised predictions by using Welch's t-test or Wilcoxon rank sum test, depending on the outcomes of the Shapiro-Wilk test of normality, also accounting for unequal sample sizes (hold-out: $N_{1}$=$10$ repetitions; LOSO-CV: $N_{2}$= $30$ or $22$ subjects).
Eventually, we combined the findings obtained from the different performance analyses (i.e., subject-independent predictions, personalised predictions, comparison of maximal performances) to confidently identify the models with superior performances, thereby enabling us to proceed with the explainability analysis using the most promising candidates.

\subsection{XAI methods}\label{sec:xai_methods}
We investigated several state-of-the-art XAI methods to enable the interpretability of the best performing ML models. Specifically, we focused on the most prominent methods used in healthcare applications to explain models learned from tabular data \cite{sahakyan2021explainable}.
These methods are generally applied to get \textit{post-hoc} explanations of already developed complex models, and especially to obtain global explanations that can be used to understand model reasoning as a whole. As a result, we selected SHAP along with Local Interpretable Model-agnostic Explanations (LIME) \cite{ribeiro2016should} and feature permutation methods.
\\
SHAP is a benchmark XAI method developed by applying the concepts of coalitional game theory to the ML domain, by considering a prediction task as a game, features as players, and coalitions as all possible feature subsets. SHAP computes feature importance scores as the average marginal contribution that each input variable brings to each prediction, and it also provides solid theoretical foundations to aggregate local outcomes into global explanations.
LIME technique is another popular model-agnostic technique for local interpretability. It generates an explanation by approximating a complex model by an interpretable surrogate model (generally, a sparse linear model) only in the neighborhood of the data instance to be explained, then it exploits the learned regression coefficients as feature importance scores. LIME works better for local interpretability, but global explanations may also be derived, for instance by computing the mean of absolute LIME weights across a given data set.
Finally, feature permutation is an additional method to estimate feature importance, by measuring how much a particular scoring metric decreases when the values of a given feature are randomly shuffled. In this way, the relationship between a feature and the model output is broken, and the drop in scoring metric is used to quantify how much the model depends on that feature. Mean Decrease in Accuracy (MDA) is a typical choice in permutation studies, but tree-based models also provide an alternative measure of feature importance known as MDI, sometimes also referred to as Gini importance. MDI is computed as the total decrease in impurity (i.e., homogeneity of labels) of leaf nodes for all the splits that involve a given feature within a tree, weighted by probability of reaching each node. This value is averaged across all learners for tree ensembles.
\\
For both datasets, all the above XAI methods have been used to compute global feature rankings for the best performing models.
Then, we conduct a per-model explanation consistency assessment by evaluating the level of agreement between explanations generated by these methods, in order to get some preliminary insights into model stability and robustness.
To this aim, for each selected model we evaluated the degree of overlap between feature rankings for each pair of XAI methods and for the top-$K$ features (with $K=$ $1$,$3$,$5$). Specifically, we measured both the number of exact matches (i.e., when the same feature is ranked at the same position) and of non-exact matches (i.e., when the same feature appears in the top-$K$ list regardless of its position).
\\
Moreover, we also propose a preliminary clinical validation of the generated explanations, as it represents a fundamental step to enhance acceptance and trust in AI-empowered decision-making for healthcare applications. To this aim, we used SHAP and LIME to investigate the global relationships between model outcomes and the most relevant predictors. Feature permutation methods do not enable revealing feature trends.
For SHAP, we used summary plots to obtain an overview of feature correlations, then we analysed individual feature trends separately using dependence plots.
In order to remove feature interactions, we computed SHAP interaction values to break down each feature trend into main effect and interaction effects, in order to focus on the main impact of each feature on model output. For LIME, weights already estimates the main effect of each feature as they represent the coefficients of the model's linear approximations. Clinical comparison has been performed both through interviews with a medical specialist, as well as through the analysis of the related medical literature.
\\
As a final remark, it should be noted that other XAI methods might also be used to explain the models, such as Anchors \cite{ribeiro2018anchors}, and Local Rule-Based Explanations of black-box decision systems (LORE) \cite{guidotti2018local}. However, these methods provide IF-THEN decision rules as explanations rather than feature attributions, while explanation consistency assessment generally applies to methods providing the same explanation format. In addition, these methods are best suited only for local explainability, therefore a consistent aggregation of local rules at global stage may represent a challenging task.

\section{Results and Discussion}
\subsection{Subject-independent predictions}
\begin{figure}[!htbp]
\begin{subfigure}{0.9\textwidth}
  \includegraphics[width=\linewidth]{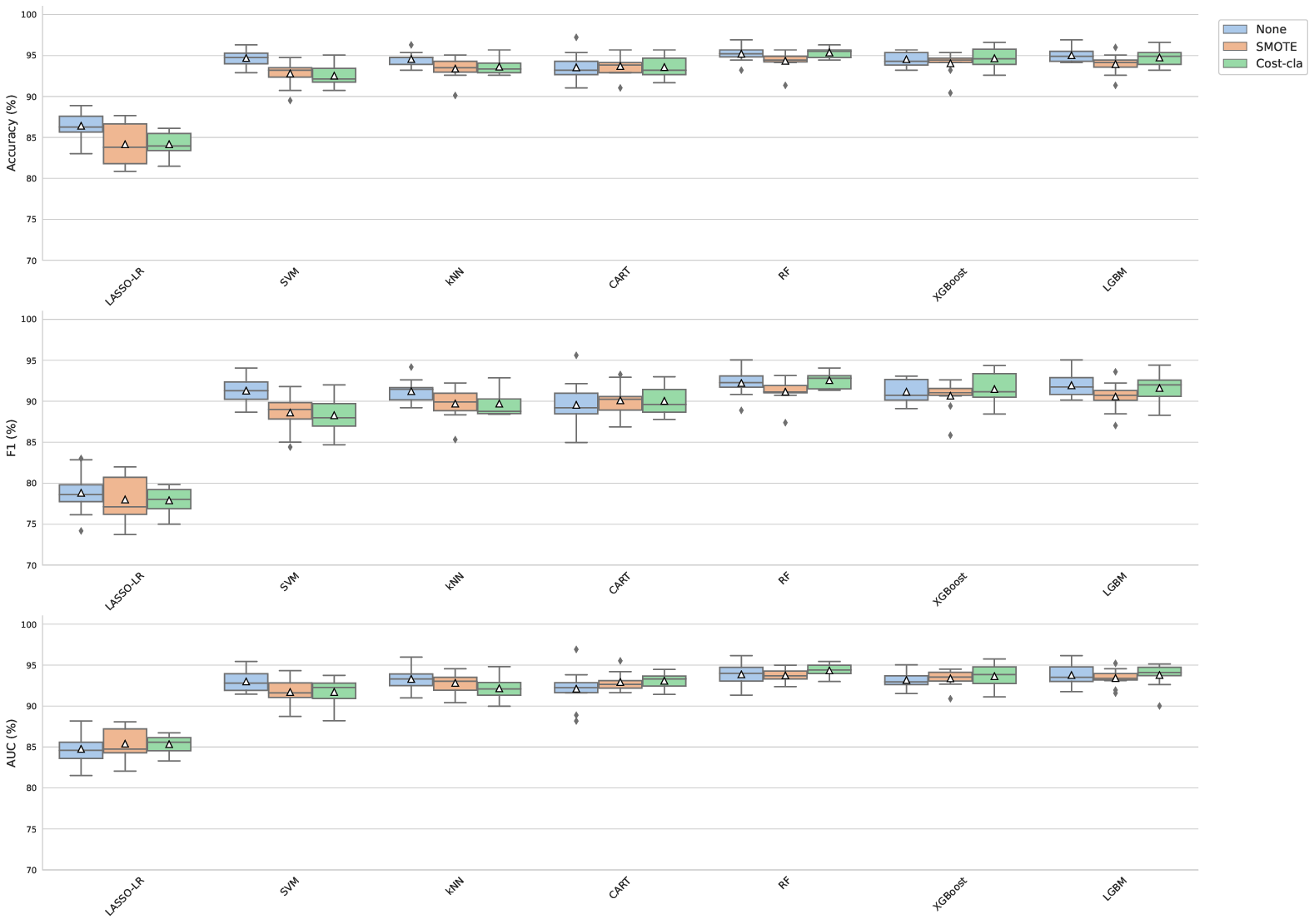}  
  \caption{With body composition data.}
  \label{fig:bodycompo_holdout_cv}
\end{subfigure}
\newline
\begin{subfigure}{0.9\textwidth}
  \includegraphics[width=\linewidth]{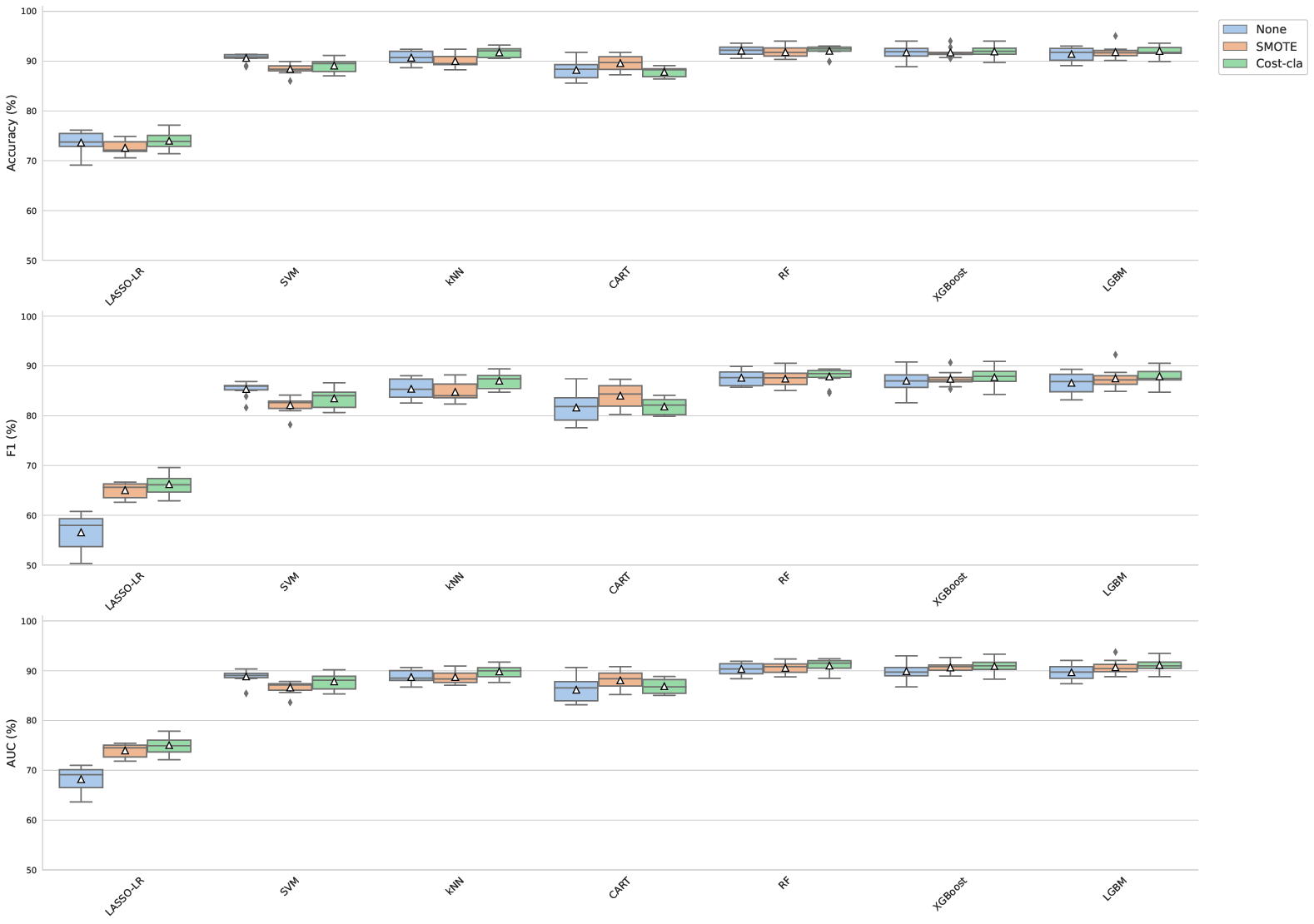}
  \caption{Without body composition data.}
  \label{fig:nutritional_holdout_cv}
\end{subfigure}
\caption{Model performances for repeated train-test splits ($N=10$ runs).}
\label{fig:holdout_cV_results}
\end{figure}

Results obtained from repeated train-test splits are shown in Figure \ref{fig:bodycompo_holdout_cv} and Figure \ref{fig:nutritional_holdout_cv}, depending on the dataset type. Evaluation metrics are grouped by the reference data imbalance management method (\textit{None} = unbalanced training data, \textit{SMOTE} = training data oversampling, \textit{Cost-cla} = cost-sensitive classifiers).
Obtained results show very narrow inter-quartile ranges (IQRs), indicating consistent performances across multiple runs with different test sets. Additionally, most distributions are non-skewed and do not present outliers, resulting in median and mean values that closely align with each other. 
Model training with original, unbalanced data generally outperform both SMOTE and cost-sensitive learning approaches for most model types, with also some cases showing a significant improvement ($\textit{p}<0.05$). Specifically, SVM accuracy and F$1$-score are higher compared to both SMOTE and cost-sensitive learning when body composition data are included (Accuracy, Cost-cla: $\textit{p}=0.0039$; Accuracy, SMOTE: $\textit{p}=0.0121$; F$1$, Cost-cla: $\textit{p}=0.0095$; F$1$, SMOTE: $\textit{p}=0.0214$).
For models without body composition data, F$1$ score and AUC are higher compared to SMOTE and cost-sensitive learning for LASSO-LR ($\textit{p}=0.001$ for all comparisons), while accuracy is higher compared to cost-sensitive learning for SVM ($\textit{p}=0.0157$). SVM performances with unbalanced training are also superior to SMOTE (accuracy: $\textit{p}=0.001$; F$1$: $\textit{p}=0.002$; AUC: $\textit{p}=0.006$).
For all other cases, $1$-way ANOVA indicate that the performance gap among the different data imbalance management methods is not statistically significant.
\\
Median accuracy of models trained in unbalanced settings ranges between $93.2\%$ (CART) to $95.2\%$ (RF) when including body composition data, whereas median F$1$ and AUC fall within $89.2\%$-$92.3\%$ and $92.2\%$-$94.0\%$, respectively. Median performances are slightly lower for models that do not include body composition assessment. Specifically, accuracy ranges between $88.4\% $ (CART) and $92.2\%$ (RF), with F$1$ between $81.3\%$ (CART) and $87.6\%$ (RF), and AUC between $84.1\%$ (SVM) and $90.3\%$ (RF). Only LASSO-LR models provide lower outcomes with respect to the other model types for both datasets.

\subsubsection{Best model analysis}
\begin{figure}[htbp]
\centering
\includegraphics[width=\textwidth]{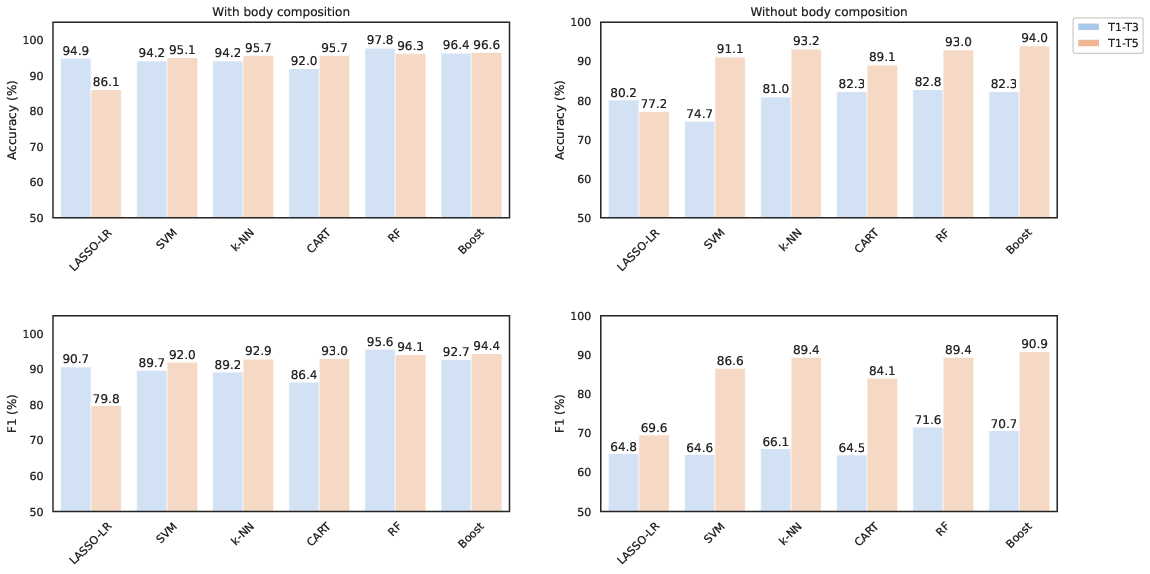}
\caption{Comparison between T$1$-T$3$ and T$1$-T$5$ best models.}\label{fig:new_VS_old_best_mdls}
\end{figure}

The comparison between the current top-performing configuration of each model type (i.e., best T$1$-T$5$ models) and those obtained in our preliminary study (i.e., best T$1$-T$3$ models) is presented in Figure \ref{fig:new_VS_old_best_mdls}. 
The comparison is based on the results obtained from the hold-out evaluation. T$1$-T$5$ models trained with unbalanced data have been selected based on the findings described in the previous section, while the best T$1$-T$3$ models are generally cost-sensitive classifiers due to the higher class imbalance in their corresponding datasets. Please also note that the \textit{Boost} label displayed on the X-axis refers to AdaBoost in the case of T$1$-T$3$ models, and to the best gradient boosting implementation (XGBoost or LightGBM) for T$1$-T$5$ models.
Models incorporating body composition data achieve maximum accuracy and F$1$ values that are comparable to those of the previous T$1$-T$3$ models for almost all model families. Specifically, there are only small differences in model accuracy, always within the range of $\pm1.7\%$.
On the other hand, most T$1$-T$5$ models exhibit slightly better F$1$ scores, ranging from $+1.7\%$ (Boost) to $+6.6\%$ (CART). This finding further confirms the high performances of models that include body composition data, indicating an improved ability to predict malnutrition risk. The only exception is the LASSO-LR model, which shows a decrease in accuracy and F$1$ scores of $-6.8\%$ and $-9.9\%$,respectively.
However, it is important to note that these results were obtained from a sample subset, as not all monitored subjects were able to undergo body composition monitoring due to physical impairments. Therefore, we attempted to extend our evaluation to a larger population without requiring body composition assessment, which resulted in a considerable improvement with respect to our previous study.
In fact, T$1$-T$5$ models combining clinical information and quantitative macro-nutrient intakes highlight significant performance gains compared to T$1$-T$3$ models based on the integration of behavioral parameters and more qualitative intake estimates. Specifically, accuracy increases from $+6.8\%$ (CART) to $+16.4\%$ (SVM), while F$1$ score increases from $+17.8\%$ (RF) to $+23.3\%$ (SVM).

\subsection{Personalised predictions}
\begin{figure}[!htbp]
\begin{subfigure}{\textwidth}
  \includegraphics[width=0.9\linewidth]{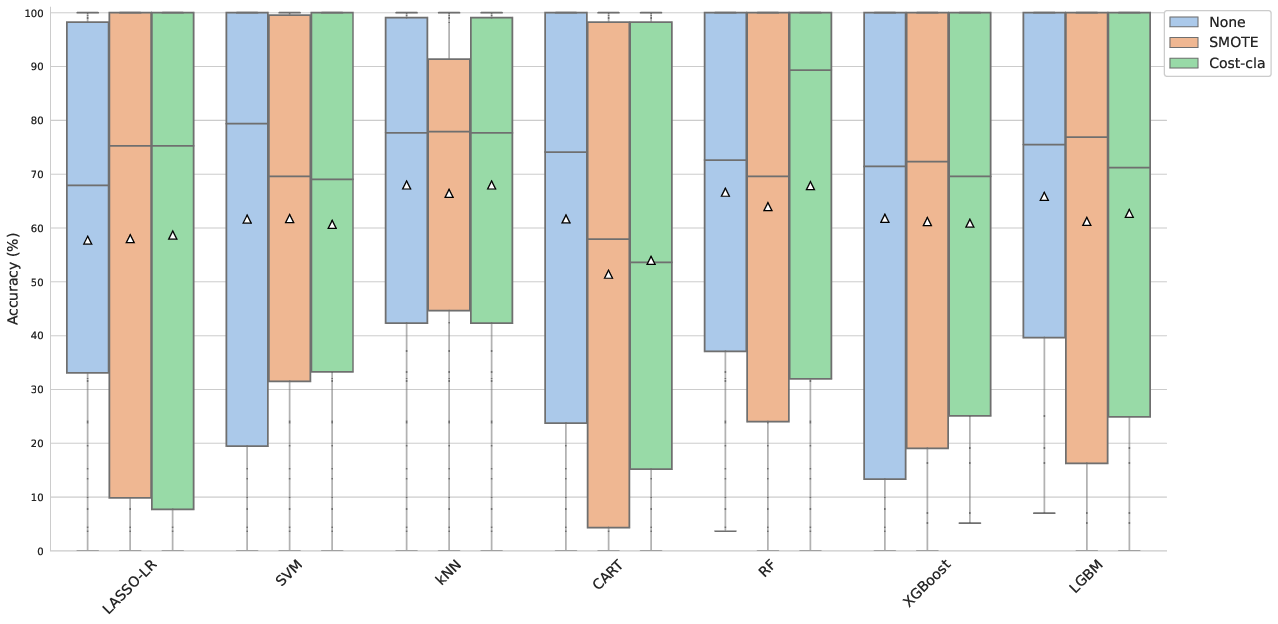}  
  \caption{With body composition data.}
  \label{fig:bodycompo_loso_cv}
\end{subfigure}
\newline
\begin{subfigure}{\textwidth}
  \includegraphics[width=0.9\linewidth]{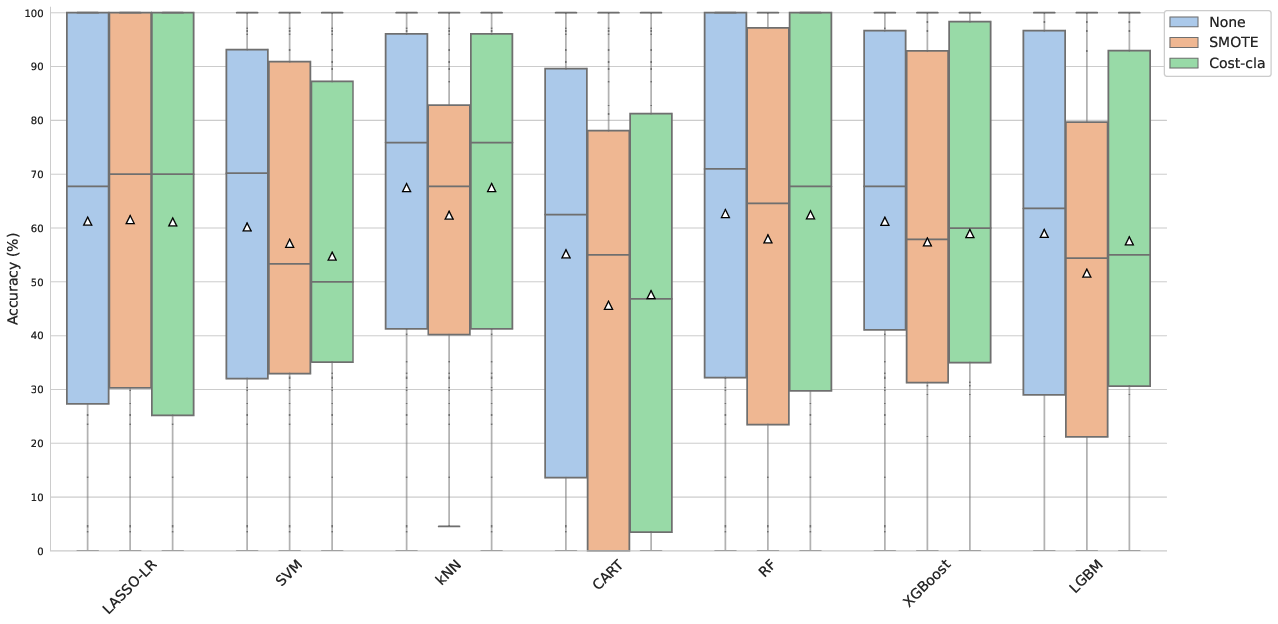}
  \caption{Without body composition data.}
  \label{fig:nutritional_loso_cv}
\end{subfigure}
\caption{Model accuracy for LOSO-CV. $N=21$ subjects with body composition, $N=30$ otherwise.}
\label{fig:loso_cv_results}
\end{figure}

Figure \ref{fig:bodycompo_loso_cv} and Figure \ref{fig:nutritional_loso_cv} present per-model accuracy obtained through LOSO-CV, with and without body composition data, respectively.
LOSO-CV results exhibit larger IQRs compared to subject-independent test sets, indicating higher variability.
Moreover, minimum accuracy values are generally zero, indicating failure cases where predictions are completely inaccurate for one or more subjects. Conversely, the $75th$ percentile values are close to or equal to $100\%$, indicating near-perfect predictions for $25\%$ of the subjects.
Models trained with unbalanced data generally yield slightly better accuracy (except for LASSO-LR), although not statistically relevant with respect to SMOTE and cost-sensitive learning as demonstrated by $1$-way ANOVA results.
When including body composition data, the best median accuracy ranges between $75.5\%$ (LightGBM) and $80\%$ (SVM), while without body composition data, it ranges between $71\%$ (SVM, RF) and $75.8\%$ (k-NN) for the best models. Notably, the median accuracy of cost-sensitive RF classifiers reaches $90\%$ when using body composition data (see Figure \ref{fig:bodycompo_loso_cv}), although the average value is in line with the other approaches.

\subsubsection{Failure case analysis}
LOSO-CV results reveal that each model fails all predictions for one or more test subjects. The number of failure cases ($N_{fail}$) varies across models, ranging between $2$ ($9.1\%$) and $3$ ($13.6\%$) when using body composition data. 
Without body composition data, the number of failures is slightly higher, ranging from $4$ ($13.3\%$) to $5$ ($16.7\%$) for all models except k-NN, which has $N_{fail}=1$. 
For both datasets, models trained with unbalanced data generally exhibit lower or equal $N_{fail}$ compared to SMOTE and cost-sensitive learning.
Among the best model types, RF and LightGBM achieve $N_{fail}=0$ when using body composition data, while XGBoost has $N_{fail}=3$. Without body composition data, these models consistently have $N_{fail}=4$, which is still lower compared to the other model types.
\\
By examining the failure cases in detail, we observed that they primarily involve the same subset of subjects in our sample. Specifically, incorrect predictions are made for subjects $\#1$, $\#13$, $\#18$, $\#30$, $\#34$, and $\#42$. Among them, subjects $\#1$, $\#34$, and $\#42$ also have body composition information available.
By looking at individual data in Table \ref{tab_subject_data}, we can immediately see that the nutritional status of these subjects has been always evaluated by the medical specialist as \textit{Normal} during the corresponding trial period(s). However, most of their clinical evaluations are very close to malnutrition threshold defined by MNA (i.e., $23.5$). In fact, they present MNA scores between $24$ and $25$. 
Therefore, the prediction failures for these specific subjects may indicate that their overall profiles, encompassing clinical, nutritional, and anthropometric features, are more similar to the malnutrition risk patterns learned from the remaining subject data.

\subsection{Comparison between model validation schemes}
The statistical comparison between LOSO-CV and hold-out results obtained with unbalanced training settings does not yield any significant difference in accuracy for each model type when including body composition data.
This trend remains consistent across the other data imbalance management methods, except for k-NN, which performs worse in making personalised predictions when SMOTE is applied ($p=0.0173$).
Likewise, the best models without body composition data do not exhibit a significant difference between subject-independent and personalised predictions. Only CART demonstrates a significantly lower accuracy in individual predictions compared to user-independent test sets ($p=0.0335$).
On the contrary, a significant drop in subject-specific accuracy is observed for all classifiers except LASSO-LR when using SMOTE. Moreover, SVM ($p=0.0228$), CART ($p=0.0228$), XGBoost ($p=0.0484$), and LightGBM ($p=0.0323$) also show a significant decrease in accuracy when employing cost-sensitive learning.

\subsection{Result summary}
Predictive performance analysis reveals that models integrating body composition data generally outperform their counterparts that do not exploit this information. Although the difference is relatively small in case of hold-out performances (Figures \ref{fig:bodycompo_holdout_cv}-\ref{fig:nutritional_holdout_cv}), a more significant decrease is observed for LOSO-CV (Figures \ref{fig:bodycompo_loso_cv}-\ref{fig:nutritional_loso_cv}), with a decrease in median accuracy of over $-10\%$.
Moreover, the obtained results strongly support the selection of models trained without any data imbalance management for reference, for both model validation schemes and both datasets. For such models, subject-independent and personalised predictions exhibit statistically comparable accuracy, although the former generally outperform the latter. Indeed, larger IQRs and the occurrence of failure cases contribute to lower median accuracy values in LOSO-CV compared to the hold-out method. As a result, the best median accuracy values decrease from $93\%$-$95\%$ to $75\%$-$80\%$ when including body composition data, and from $88\%$-$92\%$ to $71\%$-$76\%$ otherwise. Nevertheless, it should be noted that personalised predictions are $100\%$ accurate for $25\%$ of the test subjects.
\\
Both hold-out and LOSO-CV model validation methods reveal that RF and gradient boosting (LightGBM and/or XGBoost) consistently outperform other model types. These models exhibit lower failure rates in personalised predictions, and they also achieve the highest maximum performances among all models, as shown in Figure \ref{fig:new_VS_old_best_mdls}).
As a result, we selected the top-performing implementation of these models as the reference for the subsequent explainability analysis. Considering that tree ensembles are inherently complex and not self-explanatory, XAI methods are required to get \textit{post-hoc} explanations of model decisions.

\subsection{Explainability}
In this section, we report and discuss a two-step assessment of global explanations provided by different XAI methods for the best performing yet complex ML models.
We first evaluate the level of consistency between feature rankings for each model, highlighting which features are selected as the most relevant, as well as the degree of overlap between rankings generated by the different explanatory techniques. This analysis also enables to detect which XAI methods provide the highest agreement.
Finally, we also perform a preliminary clinical comparison of the generated explanations to assert that the global model reasoning adheres to well-established clinical guidelines.

\subsubsection{Explanation consistency assessment}\label{sec:exp_consistency}
\begin{figure}[htbp]
\begin{subfigure}{.5\textwidth}
  \includegraphics[height=4.5cm, width=0.9\linewidth,keepaspectratio]{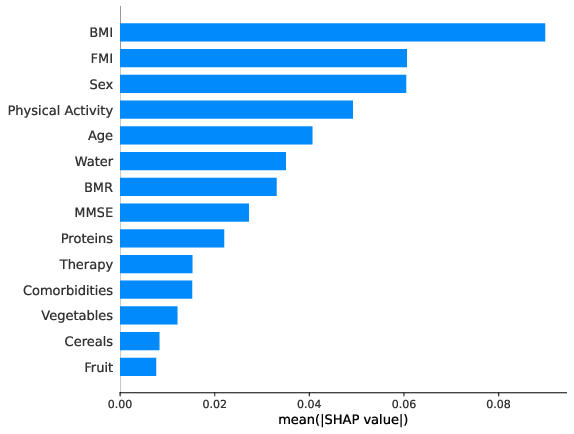}
  \caption{SHAP}
  \label{fig:shap_feat_imp_bc}
\end{subfigure}
\begin{subfigure}{.5\textwidth}
  \includegraphics[height=4.5cm, width=0.9\linewidth,keepaspectratio]{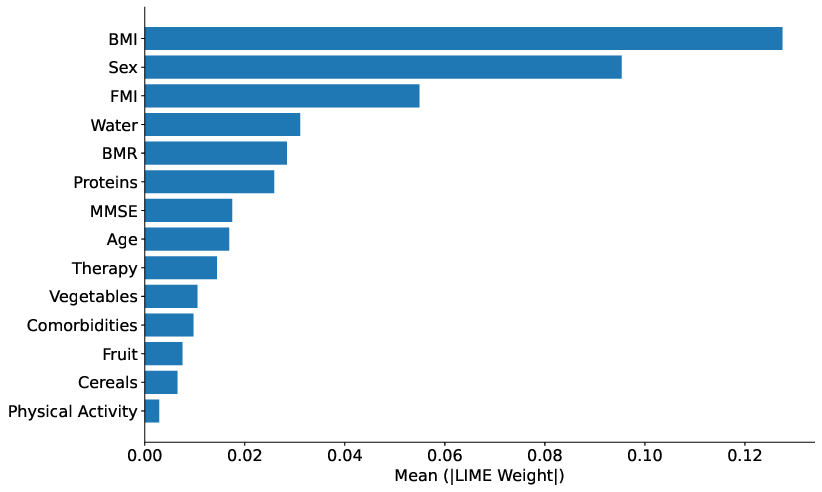}
  \caption{LIME}
  \label{fig:lime_feat_imp_bc}
\end{subfigure}
\newline
\begin{subfigure}{.5\textwidth}
  \includegraphics[height=4.5cm, width=0.9\linewidth,keepaspectratio]{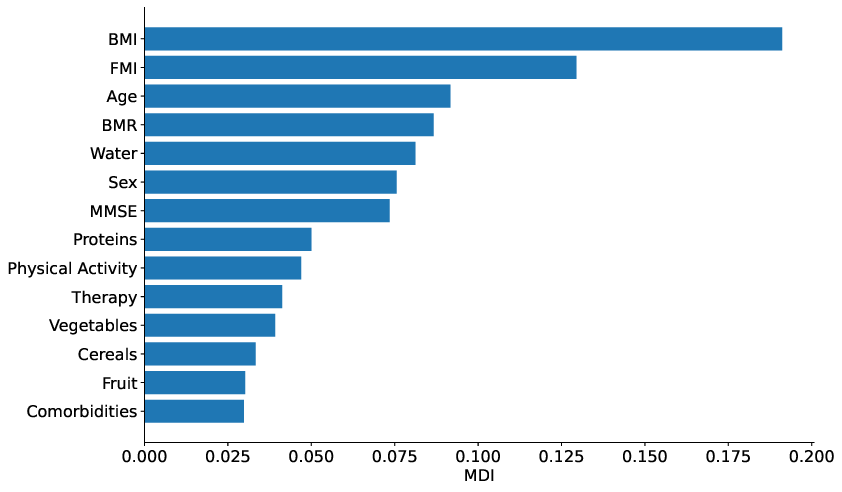}  
  \caption{MDI}
  \label{fig:mdi_feat_imp_bc}
  \end{subfigure}
  \begin{subfigure}{.5\textwidth}
  \includegraphics[height=4.5cm, width=0.9\linewidth,keepaspectratio]{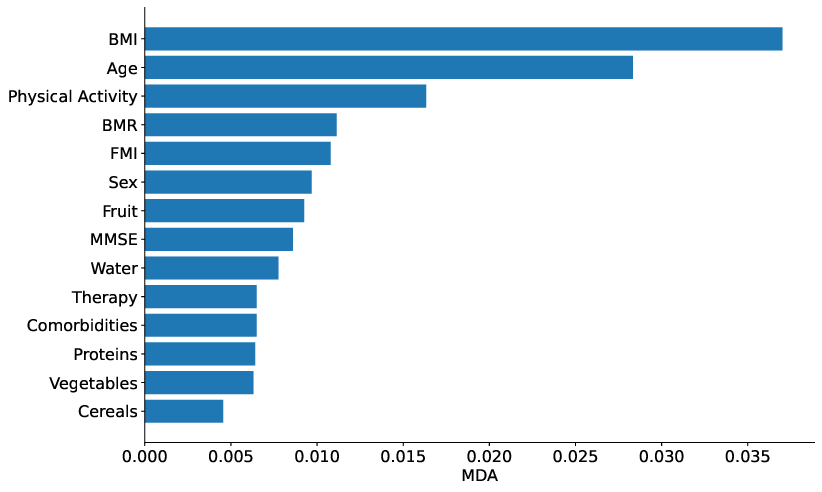}  
  \caption{MDA}
  \label{fig:mda_feat_imp_bc}
\end{subfigure}
\caption{Feature importance plots obtained for the best RF model with body composition data.}
\label{fig:RF_feat_imp_body_compo}
\end{figure}

\begin{figure}[htbp]
\begin{subfigure}{.5\textwidth}
  \includegraphics[height=4.5cm, width=0.9\linewidth,keepaspectratio]{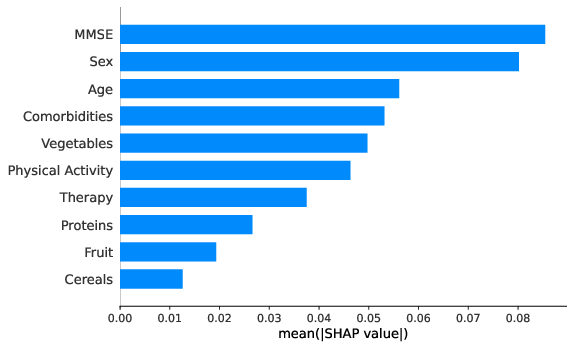}  
  \caption{SHAP}
  \label{fig:shap_feat_imp_nutrition}
\end{subfigure}
\begin{subfigure}{.5\textwidth}
  \includegraphics[height=4.5cm, width=0.9\linewidth,keepaspectratio]{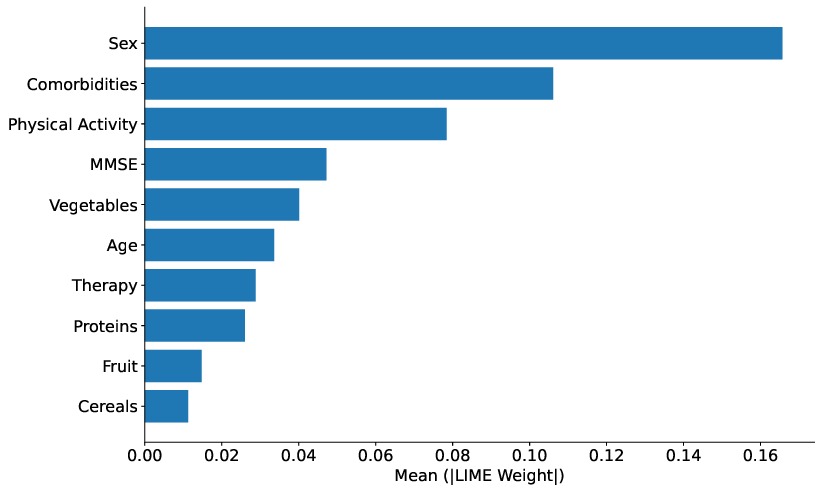}
  \caption{LIME}
  \label{fig:lime_feat_imp_nutrition}
\end{subfigure}
\newline
\begin{subfigure}{.5\textwidth}
  \includegraphics[height=4.5cm, width=0.9\linewidth,keepaspectratio]{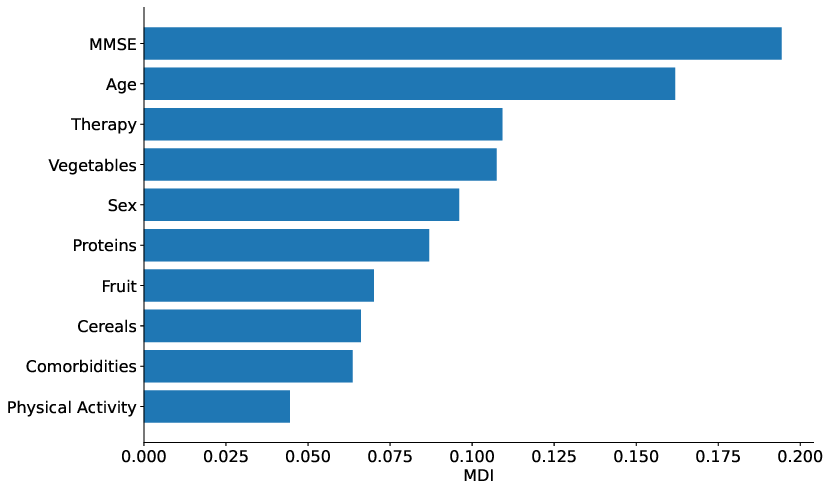}  
  \caption{MDI}
  \label{fig:mdi_feat_imp_nutrition}
  \end{subfigure}
  \begin{subfigure}{.5\textwidth}
  \includegraphics[height=4.5cm, width=0.9\linewidth,keepaspectratio]{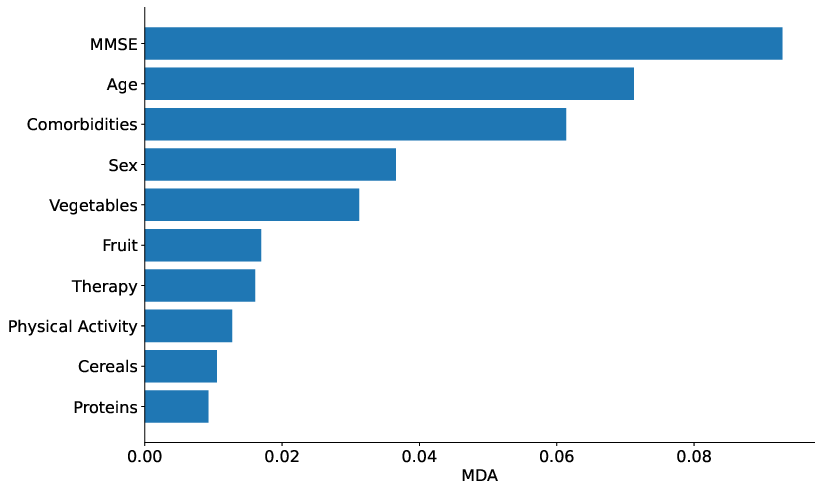}  
  \caption{MDA}
  \label{fig:mda_feat_imp_nutrition}
\end{subfigure}
\caption{Feature importance plots obtained for the best RF model without body composition data.}
\label{fig:RF_feat_imp_nutritional}
\end{figure}

\begin{figure}[!htbp]
\begin{subfigure}{.5\textwidth}
  \includegraphics[width=0.9\linewidth,]{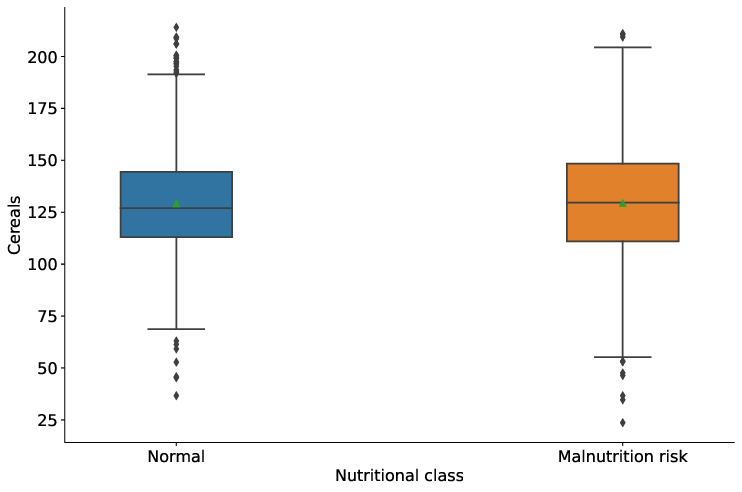}  
  \caption{Cereals}
  \label{fig:cereals}
\end{subfigure}
\begin{subfigure}{.5\textwidth}
  \includegraphics[width=0.9\linewidth]{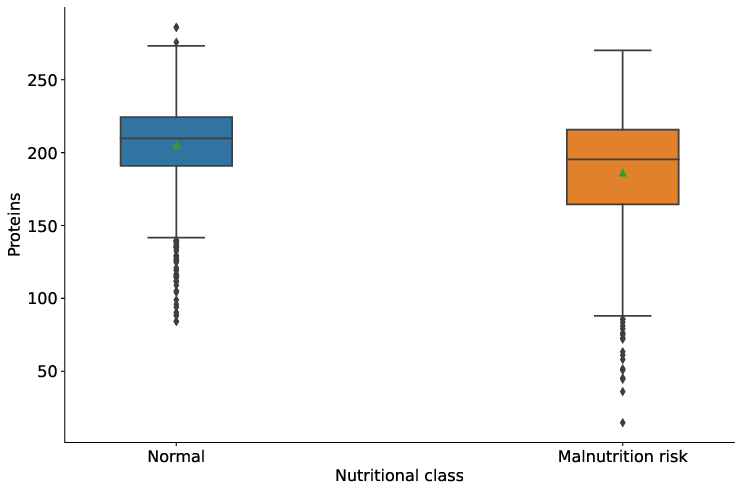}
  \caption{Proteins}
  \label{fig:proteins}
\end{subfigure}
\newline
\begin{subfigure}{.5\textwidth}
  \includegraphics[width=0.9\linewidth]{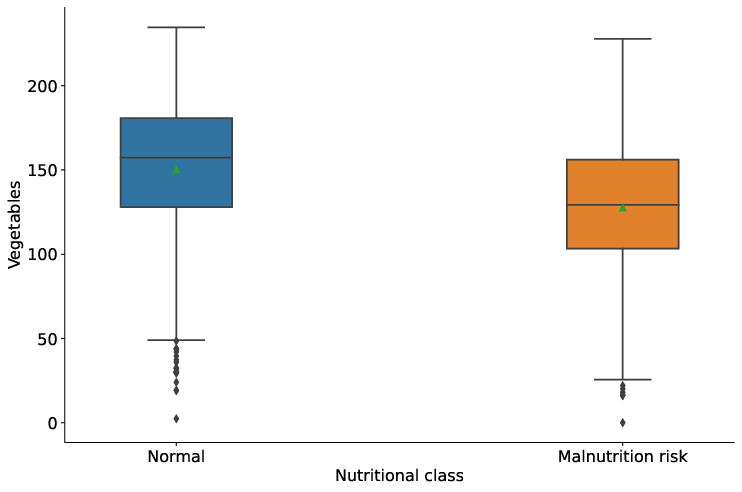}  
  \caption{Vegetables}
  \label{fig:vegetables}
  \end{subfigure}
  \begin{subfigure}{.5\textwidth}
  \includegraphics[width=0.9\linewidth]{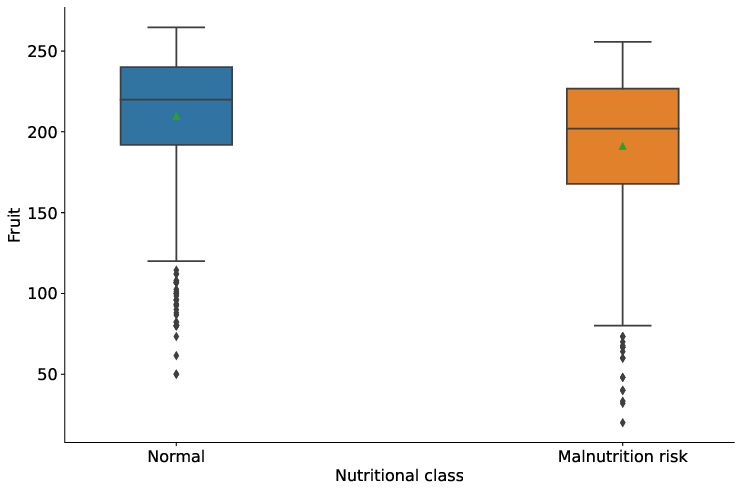}
\caption{Fruit}
\label{fig:fruit}
\end{subfigure}
\caption{Distribution of average daily intake values for each macro-nutrient among the two nutritional classes.}
\label{fig:macro-nutrient intake}
\end{figure}

Feature importance plots for the best RF models with and without body composition data are shown in Figure \ref{fig:RF_feat_imp_body_compo} and Figure \ref{fig:RF_feat_imp_nutritional}, respectively.
Metrics used for attributing feature importance by each method have been already discussed in Section \ref{sec:xai_methods}. For what concerns SHAP, model output to be explained is the prediction probability of malnutrition risk class for RF, and the log odds ratio for XGBoost/LightGBM algorithms, respectively. Therefore, SHAP ranking is computed according to the average absolute variation each feature brings to these values.
\\
In addition, we report the top-$5$ features of each ranking for XGBoost and LightGBM models in Table \ref{tab:boost_top5_feat_bodycompo} and Table \ref{tab:boost_top5_feat_bodycompo}, respectively, depending on the dataset type.
Obtained results indicate that when body composition assessment is included, the corresponding features are generally ranked as the most important, and especially in case of RF.
Almost all methods agree in ranking BMI as the most relevant predictor for all the models, with only LIME and MDI providing a different result for XGBoost and LightGBM, respectively. FMI is also one of the most important features for RF, as it is ranked second by SHAP and MDI, third by LIME, and fifth by MDA, respectively. Moreover, BMR and body water appear within the top-$5$ features in almost all RF rankings.
Differently, body composition parameters other than BMI are listed within the top-$5$ features in a lower number of cases for gradient boosting models, suggesting that these models attribute the highest importance to BMI within this group of features.
Such discrepancy between RF and XGBoost/LightGBM in allocating importance to body composition features may be due to the different way these algorithms handle multi-collinearity.
Indeed, BMI shows a high positive correlation with FMI ($\rho = 0.81$), which is obviously due to weight variations caused by body fat content. Moreover, both FMI and BMI show a moderate-to-high negative correlation with total body water ($\rho=-0.91$ and $\rho =-0.57$, respectively), since a high fat mass content usually carries much less water than the lean mass counterpart.
Finally, BMI also shows a moderate correlation with BMR ($\rho = 0.47$), which is expected as resting metabolism estimates depend on weight and height according to the Harris-Benedict equations.
In RF, each tree learner is built independently on various sub-samples randomly drawn from the training set, therefore importance may be distributed within a group of relevant yet correlated variables, but summing up the results all them may receive a high importance. Differently, boosting models are built by iteratively training tree learners on residuals; therefore, when presented with a group of correlated features, they may assign the largest weight to only one feature in the group (BMI in our case). As a result, their decision making process heavily relies on such feature.
\\
For what concerns models trained without body composition data, almost all XAI methods identify MMSE as the most relevant feature, indicating the strong influence of the cognitive status on malnutrition risk. Sex, age, and physical activity are the most relevant predictors following MMSE, listed within the top-$5$ features $10$, $9$, and $7$ times in total, respectively. Sex variable is also ranked as the most important input by LIME and MDI for XGBoost. These clinical attributes are also ranked among the most relevant predictors by models built with body composition data, generally after body composition statistics. Specifically, age variable can be found $10$ times within the top-$5$ features, whereas physical activity, MMSE, and sex occur $8$, $6$, and $6$ times, respectively.
By focusing on macro-nutrient intake estimates, it can be noticed that their global relevance is generally lower than those of body composition and clinical data, with only vegetable intake ranked within the top-$5$ features in many model rankings without body composition assessment.
This may be related to the similar distribution of average daily intake values for each food component among subjects belonging to the two nutritional classes, as shown in Figure \ref{fig:macro-nutrient intake}. Nutrient intake distributions are very similar among subjects belonging to the two classes for cereals, protein, and fruit, while the biggest drop is shown by vegetable intake for subjects at risk of malnutrition. Specifically, these subjects consume approximately $30$ grams less on average per day, which represents $40\%$ of a nominal portion of fresh/cooked vegetables (i.e., $80$ gr) provided by each daily menu. As a result, the information carried by these variables may be not sufficient to drive model predictions.
\\
In addition to analyse which features are detected as the most relevant, it is also fundamental to assess the consistency of global explanations provided by different XAI methods.
To this aim, a per-model assessment of the level of agreement between feature rankings is reported in Tables \ref{tab_consistency_body_compo} and  \ref{tab_consistency_nutritional}, divided by the reference datasets. We computed both the number of exact and non-exact matches since the gap in importance scores among consecutive variables is generally limited in each ranking.
Therefore, a high level of non-exact matches still indicates that a model mainly focuses on the same predictors to make decisions. 
For the top-$3$ features, the number of non-exact matches is $\ge2$ in $69.4\%$ of comparisons (i.e., $25$ out of $36$ total comparisons), with $3$ cases in which rankings are equal (SHAP $\cap$ LIME for RF and SHAP $\cap$ MDA for XGBoost in Table \ref{tab_consistency_body_compo}, SHAP $\cap$ MDA for XGBoost in Table \ref{tab_consistency_nutritional}).
For the top-$5$ features, the number of non-exact matches is $\ge3$ in $91.7\%$ of comparisons (i.e., $33$ out of $36$ cases). Specifically, it ranges between $3$ and $4$ in $29$ comparisons, whereas $100\%$ of agreement is achieved in $4$ cases, with SHAP and MDA detecting exactly the same top-$5$ features for all models without body composition data (see Table \ref{tab_consistency_nutritional}).
As it may be expected, the level of agreement between rankings is lower when considering exact matches. However, there are also perfect matches between feature rankings, such as SHAP $\cap$ MDI for XGBoost (Table \ref{tab_consistency_body_compo}) and SHAP $\cap$ MDA for all models (Table \ref{tab_consistency_nutritional}).
\\
By comparing each pair of XAI methods, the highest match is achieved by SHAP and MDA. This finding further supports the consistency of the generated global explanations, as they represent the most reliable and unbiased methods for estimating feature importance available in the literature. Conversely, the comparisons with LIME are generally less consistent, mainly because LIME rankings may suffer from a higher variance due to the aggregation of multiple local approximations. MDI may also have some intrinsic limitations, as it may inflate the importance of high cardinality numerical/categorical predictors. In addition, MDI importance is computed on statistics derived from training data, so it may not generalise well on unseen data, while in MDA the drop in accuracy is evaluated on the whole dataset (both training and test sets). However, in our study MDI outcomes are consistent with those provided by SHAP and MDA in most cases.
\\
The level of agreement between global explanations obtained through different XAI methods provides only a preliminary evaluation of inference stability and robustness, and additional efforts are required to draw more accurate conclusions.
For instance, sensitivity analysis is a more targeted approach to assess model weaknesses against random input perturbations and adversarials \citep{van2022comparison}.
However, existing sensitivity analysis methods mainly focus on computer vision and Natural Language Processing (NLP) tasks, while their effectiveness for models learned from tabular and time series data still has to be proven \citep{linardatos2020explainable}.
As a result, potential issues cannot be currently solved through XAI, and they are in current need of novel metrics and practices beyond extensive training and external validation to enhance inference reliability.

\begin{table}[!htbp]
\centering
\begin{tabular}{lllll}
    \toprule    \textbf{}&\textbf{SHAP}&\textbf{LIME}&\textbf{MDI}&\textbf{MDA}\\
    \cmidrule{2-5}
    \multirow{4}{*}{XGBoost}&BMI&BMI&Physical Activity&BMI\\
    \multirow{4}{*}{}&Physical Activity&Age&Sex&Physical Activity\\
    \multirow{4}{*}{}&Age&MMSE&BMI&Age\\
    \multirow{4}{*}{}&MMSE&Comorbidities&MMSE&MMSE\\
    \multirow{4}{*}{}&Sex&BMR&Age&FMI\\
    \midrule
    \multirow{4}{*}{LightGBM}&BMI&Physical Activity&BMI&BMI\\
    \multirow{4}{*}{}&Physical Activity&BMI&FMI&Age\\
    \multirow{4}{*}{}&Sex&Comorbidities&Age&Physical Activity\\
    \multirow{4}{*}{}&Age&Sex&BMR&MMSE\\
    \multirow{4}{*}{}&MMSE&FMI&MMSE&Cereals\\
    \bottomrule
\end{tabular}
\caption{Top-$5$ features for best XGBoost and LightGBM models with body composition data.}
\label{tab:boost_top5_feat_bodycompo}
\end{table}

\begin{table}[!htbp]
\centering
\begin{tabular}{lllll}
    \toprule  \textbf{}&\textbf{SHAP}&\textbf{LIME}&\textbf{MDI}&\textbf{MDA}\\
    \cmidrule{2-5}
    \multirow{4}{*}{XGBoost}&MMSE&Sex&Sex&MMMSE\\
    \multirow{4}{*}{}&Sex&Comorbidities&MMSE&Age\\
    \multirow{4}{*}{}&Age&Physical Activity&Physical Activity&Sex\\
    \multirow{4}{*}{}&Comorbidities&MMSE&Comorbidities&Comorbidities\\
    \multirow{4}{*}{}&Vegetables&Age&Therapy&Vegetables\\
    \midrule
    \multirow{4}{*}{LightGBM}&MMSE&Comorbidities&MMSE&MMSE\\
    \multirow{4}{*}{}&Sex&MMSE&Age&Age\\
    \multirow{4}{*}{}&Comorbidities&Age&Vegetables&Comorbidities\\
    \multirow{4}{*}{}&Age&Vegetables&Sex&Sex\\
    \multirow{4}{*}{}&Vegetables&Therapy&Therapy&Vegetables\\
    \bottomrule
\end{tabular}
\caption{Top-$5$ features for best XGBoost and LightGBM models without body composition data.}
\label{tab:boost_top5_feat_nutritional}
\end{table}

\begin{table}[!htbp]
\centering
\caption{Per-model comparison of the level of agreement between feature rankings for the top-$K$ features (with body composition data).}\label{tab_consistency_body_compo}%
\begin{adjustbox}{max width=\textwidth}
\begin{tabular}{cccccccccccccc}
\toprule
\multicolumn{1}{l}{\textbf{Model}}&\multicolumn{1}{l}{\textbf{K}}&\multicolumn{2}{c}{\textbf{SHAP $\cap$ LIME}}&\multicolumn{2}{c}{\textbf{SHAP $\cap$ MDI}}&\multicolumn{2}{c}{\textbf{SHAP $\cap$ MDA}}&\multicolumn{2}{c}{\textbf{LIME $\cap$ MDI}}&\multicolumn{2}{c}{\textbf{LIME $\cap$ MDA}}&\multicolumn{2}{c}{\textbf{MDI $\cap$ MDA}}\\
\cmidrule{3-4} \cmidrule{5-6} \cmidrule{7-8} \cmidrule{9-10}\cmidrule{11-12}\cmidrule{13-14}
&&\textbf{Exact}&\textbf{Not Exact}&\textbf{Exact}&\textbf{Not Exact}&\textbf{Exact}&\textbf{Not Exact}&\textbf{Exact}&\textbf{Not Exact}& \textbf{Exact}&\textbf{Not Exact}&\textbf{Exact}&\textbf{Not Exact}\\
\midrule
RF&1&1&N.A.&1&N.A.&1&N.A.&1&N.A.&1&N.A.&1&N.A.\\
RF&3&1&3&2&2&1&1&1&2&1&1&1&2\\
RF&5&1&3&2&3&2&4&1&4&1&3&2&4\\
\midrule
XGBoost&1&1&N.A.&0&N.A.&1&N.A.&0&N.A.&1&N.A.&0&N.A.\\
XGBoost&3&1&2&0&2&3&3&0&1&1&2&0&2\\
XGBoost&5&1&3&1&5&4&4&0&3&1&3&1&4\\
\midrule
LightGBM&1&0&N.A.&1&N.A.&1&N.A.&0&N.A.&0&N.A.&1&N.A.\\
LightGBM&3&0&2&1&1&1&2&0&1&0&2&1&2\\
LightGBM&5&0&3&2&3&1&4&0&2&0&2&1&3\\
\bottomrule
\end{tabular}
\end{adjustbox}
\end{table}

\begin{table}[!htbp]
\centering
\caption{Per-model comparison of the level of agreement between feature rankings for the top-$K$ features (without body composition data). }\label{tab_consistency_nutritional}%
\begin{adjustbox}{max width=\textwidth}
\begin{tabular}{cccccccccccccc}
\toprule
\multicolumn{1}{l}{\textbf{Model}}&\multicolumn{1}{l}{\textbf{K}}&\multicolumn{2}{c}{\textbf{SHAP $\cap$ LIME}}&\multicolumn{2}{c}{\textbf{SHAP $\cap$ MDI}}&\multicolumn{2}{c}{\textbf{SHAP $\cap$ MDA}}&\multicolumn{2}{c}{\textbf{LIME $\cap$ MDI}}&\multicolumn{2}{c}{\textbf{LIME $\cap$ MDA}}&\multicolumn{2}{c}{\textbf{MDI $\cap$ MDA}}\\
\cmidrule{3-4} \cmidrule{5-6} \cmidrule{7-8} \cmidrule{9-10}\cmidrule{11-12}\cmidrule{13-14}
&&\textbf{Exact}&\textbf{Not Exact}&\textbf{Exact}&\textbf{Not Exact}&\textbf{Exact}&\textbf{Not Exact}&\textbf{Exact}&\textbf{Not Exact}& \textbf{Exact}&\textbf{Not Exact}&\textbf{Exact}&\textbf{Not Exact}\\
\midrule
RF&1&0&N.A.&1&N.A.&1&N.A.&0&N.A.&0&N.A.&1&N.A.\\
RF&1&0&1&1&2&1&2&0&0&0&1&2&2\\
RF&5&0&4&1&4&1&5&0&2&0&4&2&4\\
\midrule
XGBoost&1&0&N.A.&0&N.A.&1&N.A.&1&N.A.&0&N.A.&0&N.A.\\
XGBoost&3&0&1&0&2&1&3&2&2&0&1&0&2\\
XGBoost&5&0&4&1&3&3&5&2&4&0&4&1&3\\
\midrule
LightGBM&1&0&N.A.&1&N.A.&1&N.A.&0&N.A.&0&N.A.&1&N.A.\\
LightGBM&3&0&2&1&1&2&2&0&2&0&3&2&2\\
LightGBM&5&0&4&1&4&3&5&1&4&0&4&3&4\\
\bottomrule
\end{tabular}
\end{adjustbox}
\end{table}

\subsubsection{Preliminary clinical validation}
\begin{figure}[!htbp]
\centering
\begin{subfigure}{.48\textwidth}
  \includegraphics[width=\linewidth,]{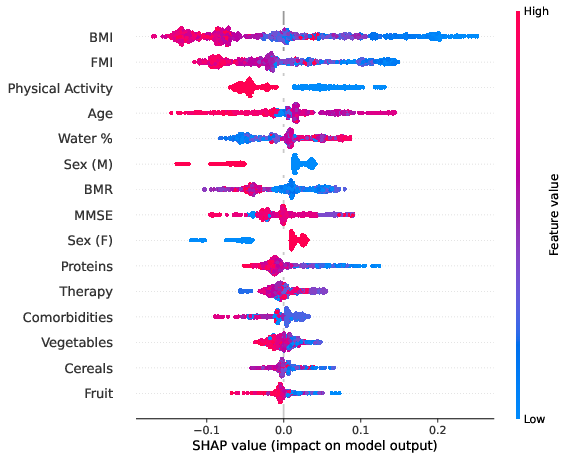}  
  \caption{With body composition data.}
  \label{fig:shap_summary_plot_bodycompo}
\end{subfigure}
\begin{subfigure}{.48\textwidth}
  \includegraphics[width=\linewidth]{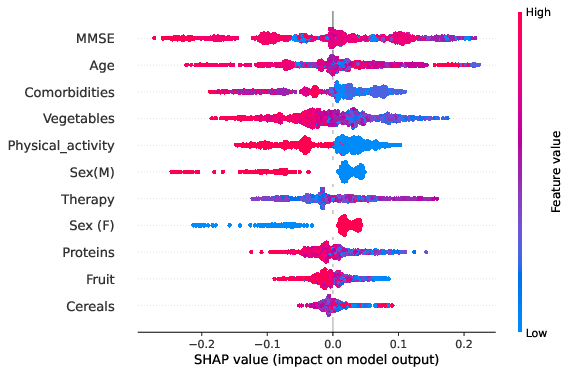}
  \caption{Without body composition data.}
  \label{fig:shap_summary_plot_nutritional}
\end{subfigure}
\caption{SHAP summary plots for the best RF models.}
\label{fig:SHAP_summary_plots}
\end{figure}

\begin{figure}[!htbp]
\begin{subfigure}{.5\textwidth}
  \includegraphics[width=0.9\linewidth,]{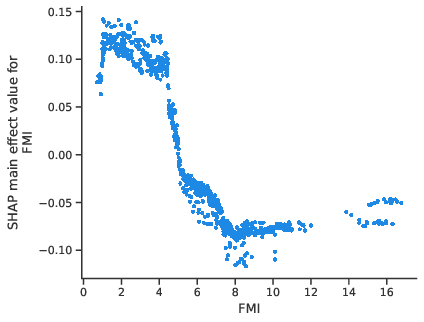}  
  \caption{FMI}
  \label{fig:shap_FMI}
\end{subfigure}
\begin{subfigure}{.5\textwidth}
  \includegraphics[width=0.9\linewidth]{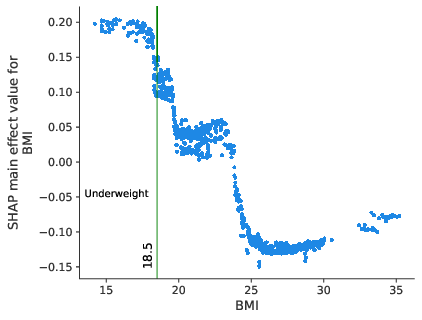}
  \caption{BMI}
  \label{fig:shap_BMI}
\end{subfigure}
\newline
\begin{subfigure}{.5\textwidth}
  \includegraphics[width=0.9\linewidth]{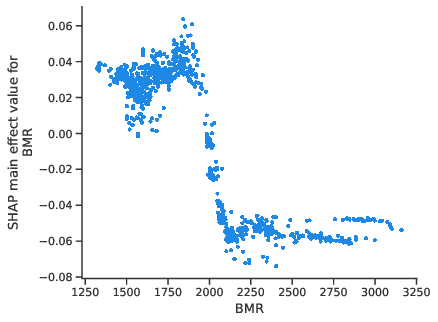}  
  \caption{BMR}
  \label{fig:shap_BMR}
  \end{subfigure}
  \begin{subfigure}{.5\textwidth}
  \includegraphics[width=0.9\linewidth]{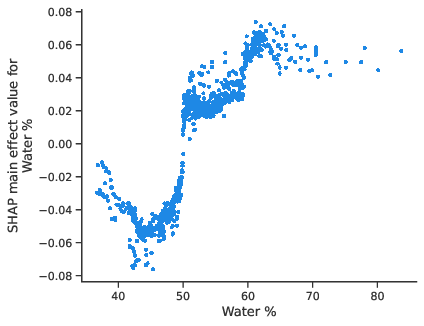}
\caption{Body water ($\%$)}
\label{fig:shap_body_water}
\end{subfigure}
\caption{SHAP main effect of body composition features on RF model output.}
\label{fig:shap_body_compo_feature_trends}
\end{figure}

\begin{figure}[!htbp]
\begin{subfigure}{.5\textwidth}
  \includegraphics[width=0.9\linewidth,]{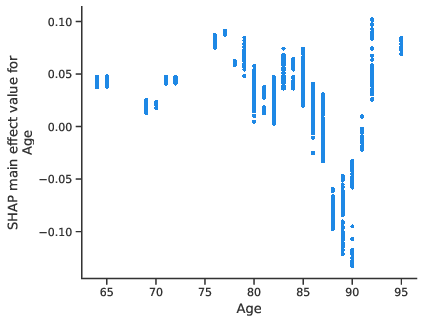}  
  \caption{Age}
  \label{fig:shap_age}
\end{subfigure}
\begin{subfigure}{.5\textwidth}
  \includegraphics[width=0.9\linewidth]{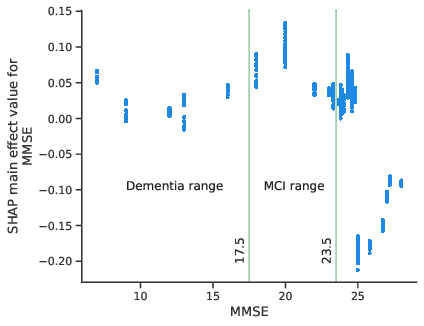}
  \caption{MMSE}
  \label{fig:shap_MMSE}
\end{subfigure}
\newline
\begin{subfigure}{.5\textwidth}
  \includegraphics[width=0.9\linewidth]{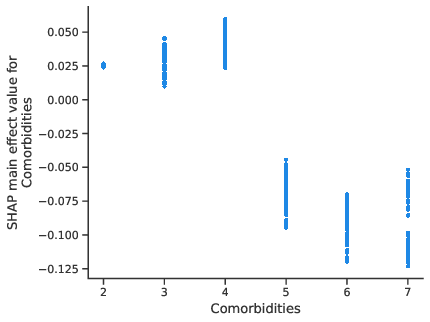}
  \caption{$\#$ of comorbidities}
  \label{fig:shap_comorbidities}
  \end{subfigure}
  \begin{subfigure}{.5\textwidth}
  \includegraphics[width=0.9\linewidth]{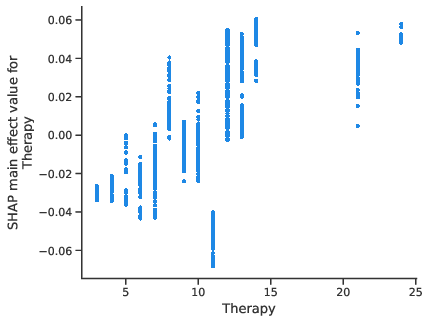}
\caption{$\#$ of therapies}
\label{fig:shap_therapy}
\end{subfigure}
\caption{SHAP main effect of clinical variables on RF model output.}
\label{fig:SHAP_clinical_feature_trends}
\end{figure}

\begin{figure}[!htbp]
\begin{subfigure}{.5\textwidth}
\centering
\includegraphics[width=\textwidth]{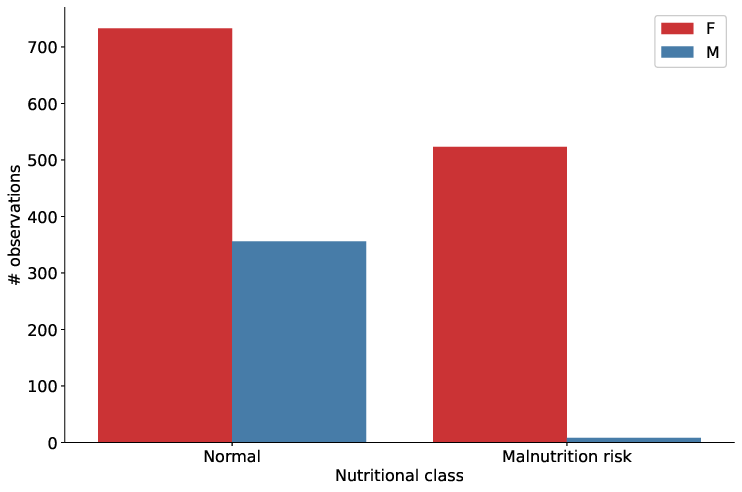}
\caption{Sex.}
\label{fig:output_vs_sex}
\end{subfigure}
\begin{subfigure}{0.5\textwidth}
\centering
\includegraphics[width=\textwidth]{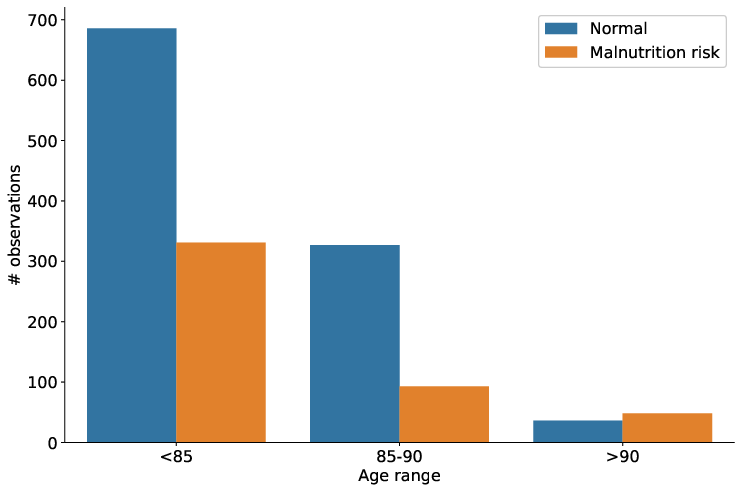}
\caption{Age range.}
\label{fig:output_vs_age_range}
\end{subfigure}
\caption{Distribution of sex and age range with respect to nutritional classes.}
\label{fig:output_vs_age_and_sex}
\end{figure}

\begin{figure}[!htbp]
\begin{subfigure}{.5\textwidth}
  \includegraphics[width=0.9\linewidth,]{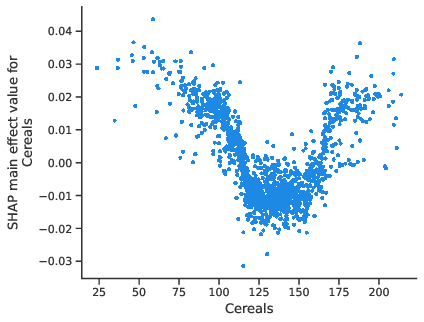}
  \caption{Cereals}
  \label{fig:shap_cereal_trend}
\end{subfigure}
\begin{subfigure}{.5\textwidth}
  \includegraphics[width=0.9\linewidth]{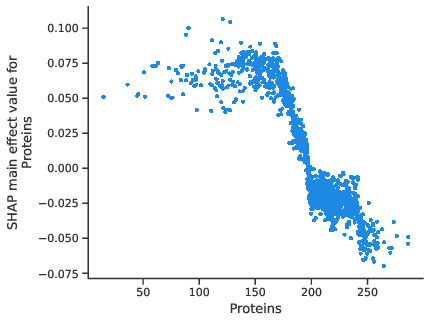}
  \caption{Proteins}
  \label{fig:shap_protein_trend}
\end{subfigure}
\newline
\begin{subfigure}{.5\textwidth}
  \includegraphics[width=0.9\linewidth]{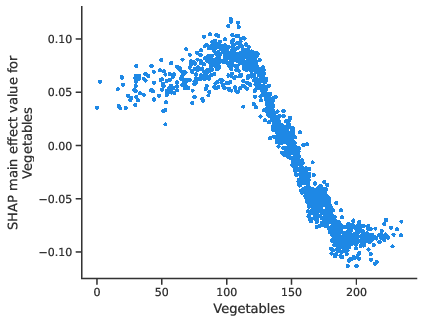}  
  \caption{Vegetables}
  \label{fig:shap_vegetable_trend}
  \end{subfigure}
  \begin{subfigure}{.5\textwidth}
  \includegraphics[width=0.9\linewidth]{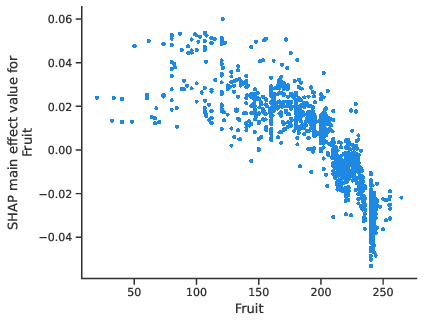}
\caption{Fruit}
\label{fig:shap_fruit_trend}
\end{subfigure}
\caption{SHAP main effect of average daily intake estimates of macro-nutrients on RF model output.}
\label{fig:shap_macro_nutrient_feature_trends}
\end{figure}

SHAP summary plots obtained for the best RF models are shown in Figure \ref{fig:SHAP_summary_plots}, divided by the reference dataset. In addition, summary plots obtained through LIME are reported in Figure \ref{fig:lime_summary_plot_bodycompo} and Figure \ref{fig:lime_summary_plot_nutritional} in the Appendix, respectively.
In these plots, feature are still ranked in descending order of importance, but the X-axis also shows the importance score associated with every observation in the dataset.
Moreover, the colormap indicates whether the value of a feature is high or low for the corresponding observation, in order to enable discovering input-output relationships.
Please note that the impact of sex variable is split for each category separately since this variable cannot be scaled into high/low values.
From summary plots, it is possible to immediately detect some global correlations, which are supported by looking at individual feature dependence plots. 
The FMI trend (Figure \ref{fig:shap_FMI}) matches with the negative correlation with malnutrition risk known in the clinical practice, confirming a strong association between fat mass depletion and malnutrition risk \citep{tomlinson2019body}. 
A clearly negative correlation is also shown between BMI and malnutrition risk in Figure \ref{fig:shap_BMI}.
Most remarkably, the highest positive impact on malnutrition risk probability is shown for BMI values below $20$, very close to the clinical threshold of $18.5$ used to distinguish between underweight and normal weight.
Specifically, subjects having BMI $<18.5$ show an increase in the malnutrition risk prediction probability of approximately $+20\%$. Then, impact scores rapidly decrease within the normal weight range ($18.5\le$BMI$\ge25$), reaching negative contributions up to $-15\%$ in the overweight range (BMI $>25$). As a result, the overall BMI trend confirms underweight condition as one of the main indicators for malnutrition risk assessment \citep{norman2021malnutrition}.
BMR also exhibits a negative correlation with malnutrition risk, as shown in Figure \ref{fig:shap_BMR}. Positive scores are associated with BMR values approximately below $1850$ KCal/day, then they rapidly decrease becoming negative for BMR $\ge2000$ KCal/day.
BMR is generally higher for men than women, and it decreases almost linearly with normal ageing, mainly due to the loss of skeletal musculature that is responsible for consuming the largest amount of energy in the human body. However, with prolonged underfeeding BMR is further reduced to adapt to chronic energy deficiency, and especially to suppress gluconeogenic activity in order to minimise muscle protein breakdown and avoid ketones becoming the main substrates for energy supply to the brain \citep{emery2005metabolic}.
As a result, the resting metabolism tends to further decrease in older subjects exposed to an extended malnutrition risk condition, in line with the negative trend learned by the model.
Differently, total body water percentage shows a positive correlation with malnutrition risk (see Figure \ref{fig:shap_body_water}).
This value includes both intracellular water (ICW) and extracellular water (ECW) content, and it manifests a linear decrease with normal ageing mainly due to an increasing ratio between fat and fat-free mass, until it constitutes less than $50\%$ of body weight in very old subjects \citep{powers2012total}.
Moreover, dehydration is a common condition in the older population \citep{paulis2018prevalence}, due to low water and water-rich food intake (i.e., vegetables and fruit) and/or to underlying illnesses (e.g., gastrointestinal disorders), which may result in a reduction of ICW \citep{lauriola2018neurocognitive}.
However, several research works have demonstrated that total body water is often increased in malnourished subjects because of high extracellular fluid retention \citep{girma2016bioimpedance}, often manifested as oedema. Indeed, the ratio of extracellular/total body water has been proposed as indicators of water balance in the context of nutritional assessment in older adults \citep{malczyk2016body}.
As a result, the observed positive trend in total body water is in line with the current clinical findings, mainly due the high impact of extracellular water accumulation.
\\
By analysing clinical variables, an expected result is that physically inactivity, due for instance to declining mobility or functional disability, provides a positive contribution to the onset and worsening of malnutrition risk among older adults \cite{cederholm2014role}.
Moreover, women are associated with a higher risk of malnutrition with respect to men, also due to the intrinsic characteristic of supplied data. Indeed, approximately $77\%$ of observations in our dataset comes from women, with an unbalanced distribution among the nutritional classes.
As shown in Figure \ref{fig:output_vs_sex}, the ratio between women and men is approximately $2$:$1$ for subjects in a normal nutritional status, while almost all subjects at risk of malnutrition are women (there is only a negligible amount of observations taken by one old male subject).
This situation is in line with the higher prevalence of malnutrition in women \citep{grammatikopoulou2019food}.
Some studies suggest that being female is one of the main risk factors for developing nutritional deficiencies \citep{dewan2008malnutrition} and altered body composition with respect to healthy, age-related standards \citep{larburu2022key,santoro2018cross}.
In addition it is known that biological, behavioral and environmental factors contribute to increase life expectancy in women \citep{baum2021new}, therefore posing them to higher chances of developing malnutrition with the progression of ageing. This general situation is even more pronounced in LTC facilities, as women are normally institutionalised at advanced age and with enhanced frailty conditions, therefore they are more likely to develop malnutrition over the time.
\\
SHAP feature trends for the other clinical attributes are shown in Figure \ref{fig:SHAP_clinical_feature_trends}, considering the largest pool of subjects (i.e., without body composition assessment) as reference.
For what concerns age variable, its impact on malnutrition risk probability is positive with a slightly increasing trend until a threshold of approximately $85$ years old, after which the trend is flipped showing a negative impact on prediction probability within the age range $86$-$90$ (see Figure \ref{fig:shap_age}). Negative importance scores are also shown in the corresponding LIME feature trend for age values $\ge85$ years (see Figure \ref{fig:lime_age} in the Appendix). Then, the impact rapidly increases reaching again a positive contribution until the maximum observed age value ($95$ years).
Age represents an independent risk factor for malnutrition, especially for subjects aged above $72$ years \citep{grammatikopoulou2019food}.
Therefore, the partially positive relationship between age and malnutrition risk shown for age values is aligned with clinical knowledge, while the drop in importance shown between $86$ and $90$ years may seem a counter-intuitive result at a first sight.
However, a possible explanation to this behaviour may be related to the occurrence of a cluster of subjects with a healthy nutritional status in such age range within our subject sample.
As shown in Figure \ref{fig:output_vs_age_range}, the class ratio is approximately $2.0$ for subjects aged $<85$ years old, whereas it increases up to $3.7$ within the range $86$-$90$.
As a result, the presence of this cluster may alter age contribution in such range.
From a clinical standpoint, this cluster might be linked to the onset of a process of homeostasis in very old adults (age $\ge85$ years), which leads to preserve their overall wellbeing status (and a good nutritional status as well) at advanced age.
However, further long-term monitoring of a larger group of subjects is required to safely generalise this finding to our target population.
\\
The MMSE trend supports the strong association between cognitive abilities and malnutrition risk. The impact of MMSE on malnutrition risk probability is positive for subjects suffering from dementia or MCI (i.e., scores $<24$). In particular, the highest scores are shown within the MCI range (up to $+15\%$), whereas the contribution is lower in the dementia range ($\le+5\%$). Then, a rapid drop occurs at the threshold between cognitively impaired and cognitively healthy subjects, resulting in a negative impact on malnutrition risk for higher MMSE values.
Both MCI and dementia conditions are known as determinants of malnutrition in the older population \citep{kimura2019malnutrition}; however, dementia leads to the greatest deterioration of dietary habits if no nutritional guidance is provided to the subjects.
However, food selection in LTC facilities is normally mediated by the nursing care personnel in case of subjects suffering from dementia, therefore the effect of unhealthy dietary habits due to severe cognitive deficits may be mitigated. On the other hand, MCI subjects have higher freedom in selecting food preferences from daily menus, which causes a higher impact of wrong behaviours on their nutritional status over the time.
\\
The coexistence of chronic comorbidities and relative pharmacotherapy, typical of frail older adults, is frequently associated with a malnourished status or with a higher risk to develop malnutrition \citep{fjell2018risk,van2016factors}.
As far as the number of therapies is concerned, Figure \ref{fig:shap_therapy} shows a global positive association between the amount of pharmacological treatments and the malnutrition risk probability, even if with a high variance in importance scores between observations sharing similar/equal values (i.e., vertical dispersion of SHAP values).
On the other hand, from Figure \ref{fig:shap_comorbidities}) we can see that a number of comorbidities $\le4$ is associated to a positive impact on malnutrition prediction probability, whereas higher values are associated to negative contributions.
However, the range of chronic disorders observed in our pool of subjects is limited (from $2$ to $7$), and this prevents us from drawing more accurate and general conclusions.
Moreover, it should be acknowledged that simply summing the number of chronic disorders and pharmacological treatments do not account for their different individual impact on the user nutritional status. For instance, diabetes, cardiovascular and chronic obstructive pulmonary diseases generally provide a major negative contribution, inducing anorexia, increased protein catabolism and impaired immune functions \citep{norman2021malnutrition,aleman2008prevalence}.
On the other hand, hypertension and hypercholesterolemia may play a less relevant role in malnutrition as they can be more easily controlled through a proper diet and/or targeted pharmacological treatments \citep{ferri2017management}.
Moreover, specific pharmacological treatments such as those for hypertension, may also benefits chronic cardiac and cerebral diseases, providing positive effects also to the nutritional status \citep{yandrapalli2019drug}, while other treatments may worsen it. From a general standpoint, our findings suggest that the categorisation of malnutrition-related diseases and drug classes is necessary to detect relevant associations.
\\
Eventually, the feature trends related to nutrient intake estimates reported in Figure \ref{fig:shap_macro_nutrient_feature_trends} indicate that the RF model detected meaningful correlations between malnutrition risk and each macro-nutrient. Specifically, the most enhanced and (almost) linear negative trends are shown for vegetables (Figure \ref{fig:shap_vegetable_trend}) and animal proteins (Figure \ref{fig:shap_protein_trend}), since chronic deficiencies in animal/vegetal proteins and micro-nutrients (i.e., vitamins and mineral salts) are the primary causes of protein-energy undernutrition \citep{van2014determinants}.
On the other hand, cereals trend shown in Figure \ref{fig:shap_cereal_trend} highlights that both carbohydrate-rich food under- and over-intake negatively impact on the user nutritional status in a similar degree.
\\
As closing remark, it is important to also highlight that LIME feature trends reported in the Appendix for body composition (Figure \ref{fig:LIME_body_compo_feature_trends}), clinical (Figure \ref{fig:LIME_clinical_feature_trends}), and nutritional (Figure \ref{fig:LIME_nutritional_feature_trends}) features are in line with those obtained by SHAP. Moreover, RF feature trends generally match with those learned by XGBoost and LightGBM models.
\\
The proposed preliminary clinical validation demonstrates that the learned input-target relationships for the most relevant predictors are clinically aligned, considering also the underlying characteristics of supplied data. 
Moreover, the best performing models do not exhibit any severe erroneous or counter-intuitive assumption, namely AI bias, which represents a fundamental aspect when AI systems are intended to go through certification and approval processes as it may save time, resources, and user efforts. 

\section{Conclusions}
The integration of m-health and XAI has the potentiality to provide effective and acceptable solutions in the healthcare domain and the malnutrition risk is an important example of a real application.
In this paper we propose a framework that integrates clinical information, heterogeneous m-health monitoring data, ML algorithms, and XAI methods in order to achieve explainable and early malnutrition risk prediction in institutionalised older adults, which may be further adapted to independent living scenarios.
Predictive performance analysis shows that the best performing models achieve accuracy and F$1$-score far above $90\%$ with subject-independent test sets, whereas personalised model predictions reach $80\%$ median accuracy with the support of body composition data, with also $100\%$ accuracy for $1$ out of $4$ test subjects. The obtained results also highlight significant performance gains with respect to our first pilot study.
Moreover, the comparison of global explanations obtained through several benchmark XAI methods indicate that the best performing models also privilege the same subset of relevant predictors to drive their decisions. This information may help AI experts to technically validate the model, while local explanations need to be integrated for model consumers (i.e., clinicians) to understand single/group predictions of interest.
Finally, we conduced a preliminary clinical validation that demonstrates that the input-target relationships learned by the models for the most relevant predictors are clinically relevant. Therefore, the global reasoning of the best performing models can be considered \textit{\enquote{human-like}} to a large extent, increasing their clinical credibility.
\\
The current framework may be further refined in order to improve both malnutrition risk prediction and explainability.
Additional long-term monitoring data can increase model inference reliability, whereas the inclusion of additional variables such as micro-nutrients, specific drug classes and chronic diseases may potentially reveal strong, clinically relevant associations with the malnutrition risk condition.
However, based on the promising classification performances and the consistency and clinical soundness of model explanations, we believe that the proposed AI-empowered m-health system may provide a valuable support to the standard clinical malnutrition assessment in LTC settings.
Moreover, explainable and early subject detection is an essential requirement to timely plan personalised malnutrition prevention and recovery strategies, such as nutritional coaching and/or pharmacological treatments, and to monitor their effectiveness over the time.

\section*{Acknowledgments}
This work has been partially funded by the European Commission under H2020-INFRAIA-2019-1SoBigData-PlusPlus project. Grant No.: 871042

\bibliographystyle{model5-names}\biboptions{authoryear}
\bibliography{main.bib}

\appendix
\pagebreak
\section{LIME results}\label{app_A}
\begin{figure}[!htbp]
\centering
\begin{subfigure}{.48\textwidth}
  \includegraphics[width=\linewidth,]{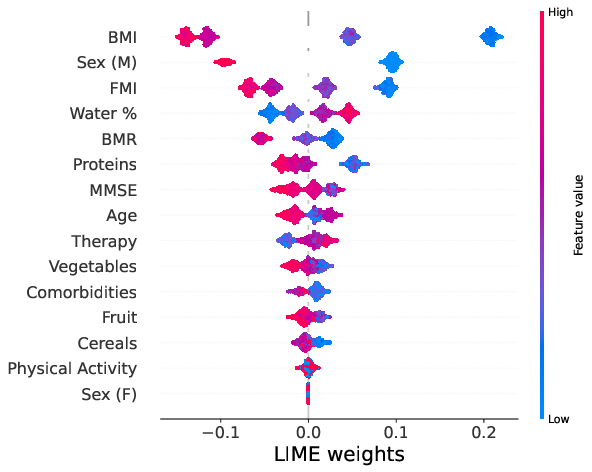}  
  \caption{With body composition data.}
  \label{fig:lime_summary_plot_bodycompo}
\end{subfigure}
\begin{subfigure}{.48\textwidth}
  \includegraphics[width=\linewidth]{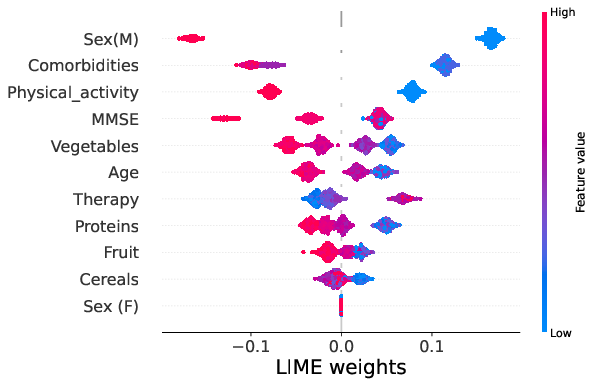}
  \caption{Without body composition data.}
  \label{fig:lime_summary_plot_nutritional}
\end{subfigure}
\caption{LIME summary plots for the best RF models.}
\label{fig:LIME_summary_plots}
\end{figure}

\begin{figure}[!htbp]
\begin{subfigure}{.5\textwidth}
  \includegraphics[width=0.9\linewidth,]{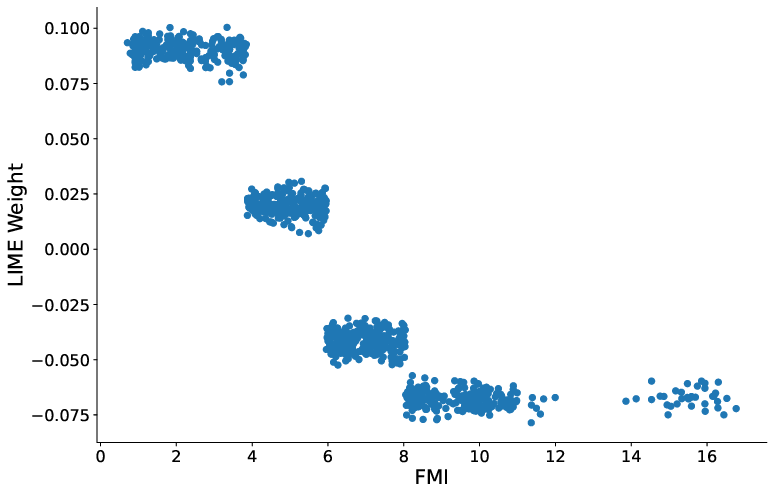}  
  \caption{FMI}
  \label{fig:lime_FMI}
\end{subfigure}
\begin{subfigure}{.5\textwidth}
  \includegraphics[width=0.9\linewidth]{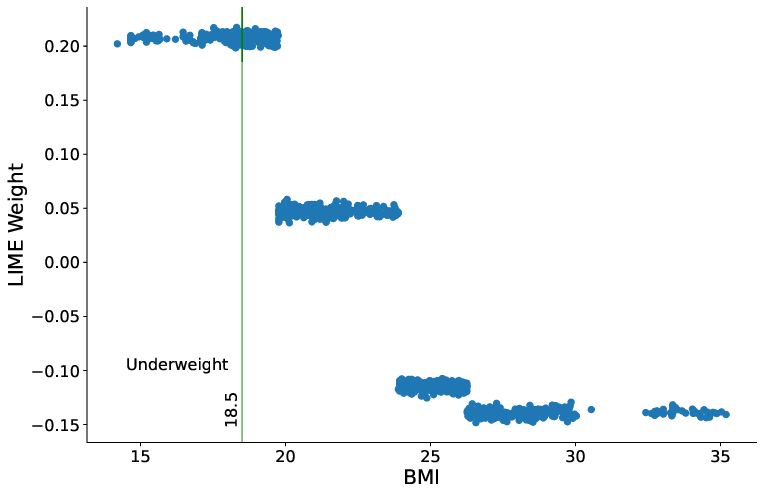}
  \caption{BMI}
  \label{fig:lime_BMI}
\end{subfigure}
\newline
\begin{subfigure}{.5\textwidth}
  \includegraphics[width=0.9\linewidth]{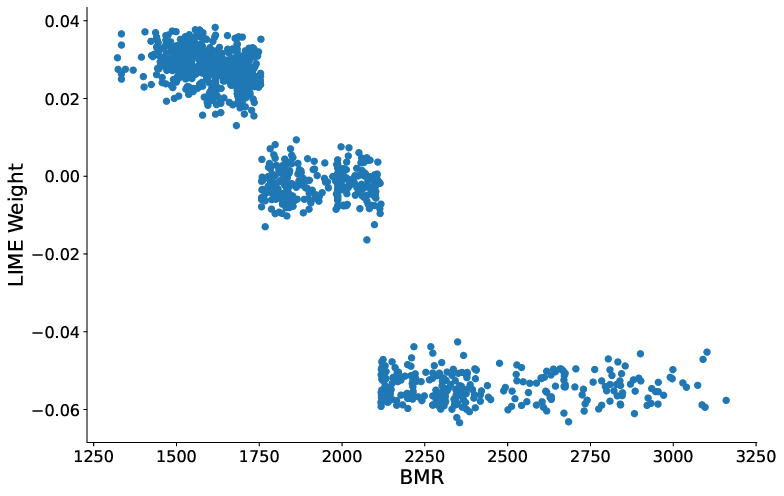}  
  \caption{BMR}
  \label{fig:lime_BMR}
  \end{subfigure}
  \begin{subfigure}{.5\textwidth}
  \includegraphics[width=0.9\linewidth]{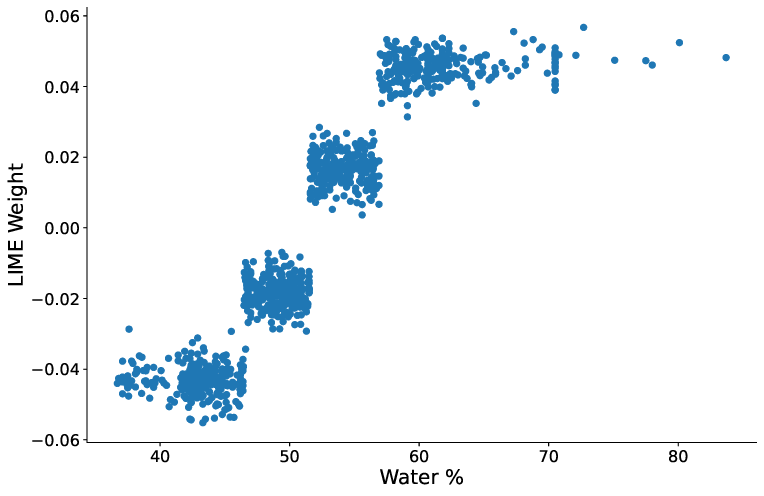}
\caption{Body water ($\%$)}
\label{fig:lime_body_water}
\end{subfigure}
\caption{LIME body composition feature trends for RF model.}
\label{fig:LIME_body_compo_feature_trends}
\end{figure}

\begin{figure}[htbp]
\begin{subfigure}{.5\textwidth}
  \includegraphics[width=0.9\linewidth,]{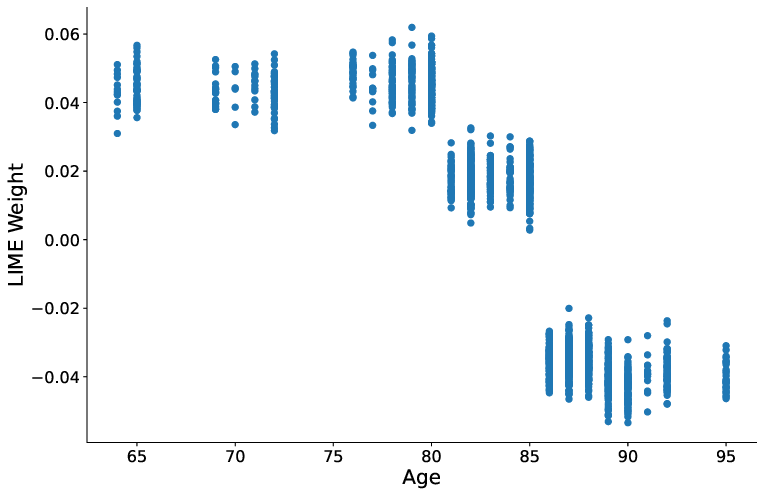}  
  \caption{Age}
  \label{fig:lime_age}
\end{subfigure}
\begin{subfigure}{.5\textwidth}
  \includegraphics[width=0.9\linewidth]{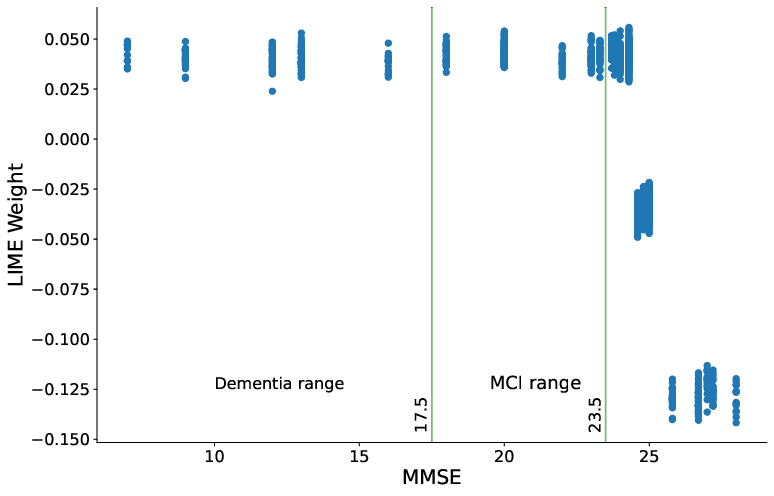}
  \caption{MMSE}
  \label{fig:lime_MMSE}
\end{subfigure}
\newline
\begin{subfigure}{.5\textwidth}
  \includegraphics[width=0.9\linewidth]{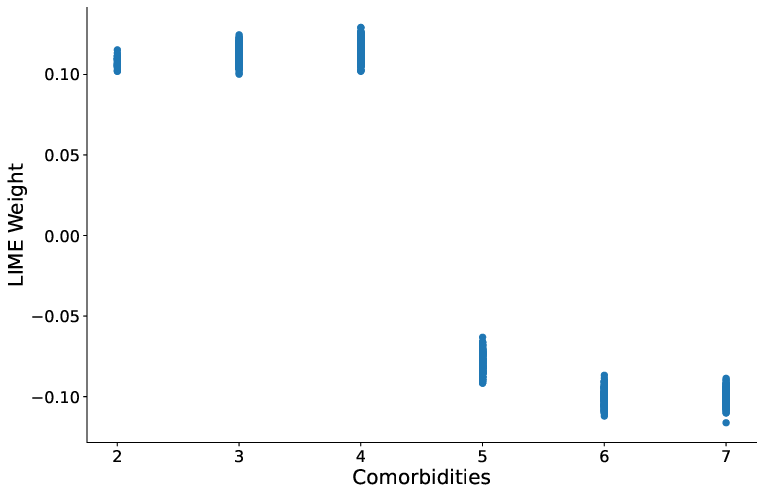}  
  \caption{$\#$ of comorbidities}
  \label{fig:lime_comorbidities}
  \end{subfigure}
  \begin{subfigure}{.5\textwidth}
  \includegraphics[width=0.9\linewidth]{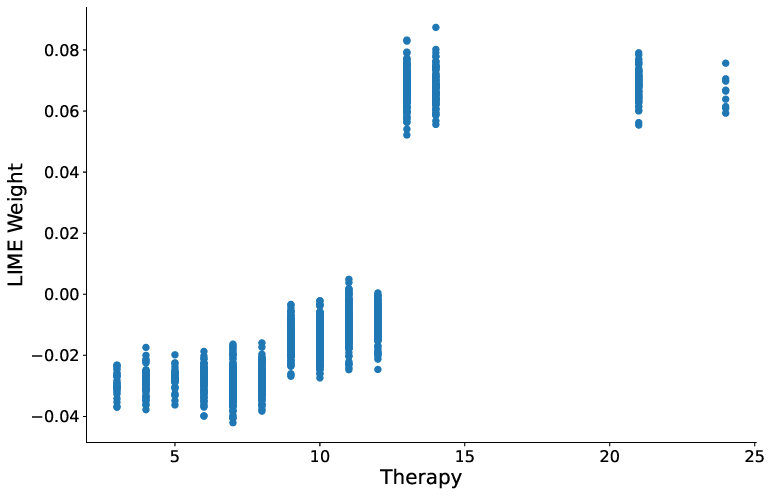}
\caption{$\#$ of therapies}
\label{fig:lime_therapy}
\end{subfigure}
\caption{LIME clinical feature trends for RF model.}
\label{fig:LIME_clinical_feature_trends}
\end{figure}

\begin{figure}[!htbp]
\begin{subfigure}{.5\textwidth}
  \includegraphics[width=0.9\linewidth,]{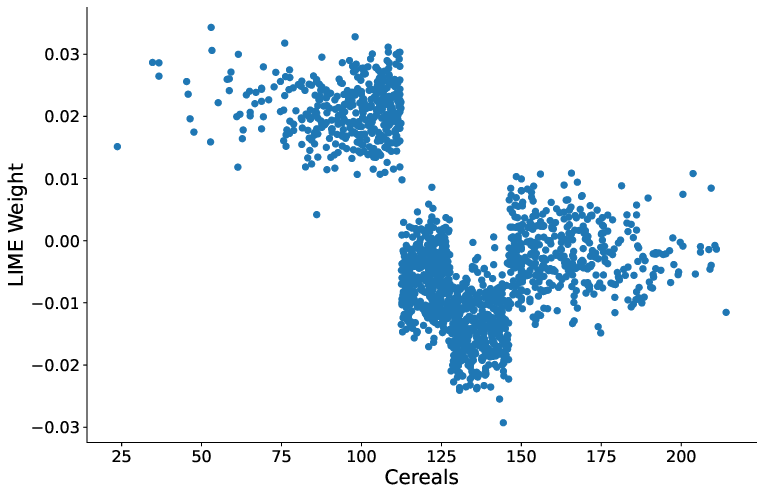}
  \caption{Cereals}
  \label{fig:lime_cereal_trend}
\end{subfigure}
\begin{subfigure}{.5\textwidth}
  \includegraphics[width=0.9\linewidth]{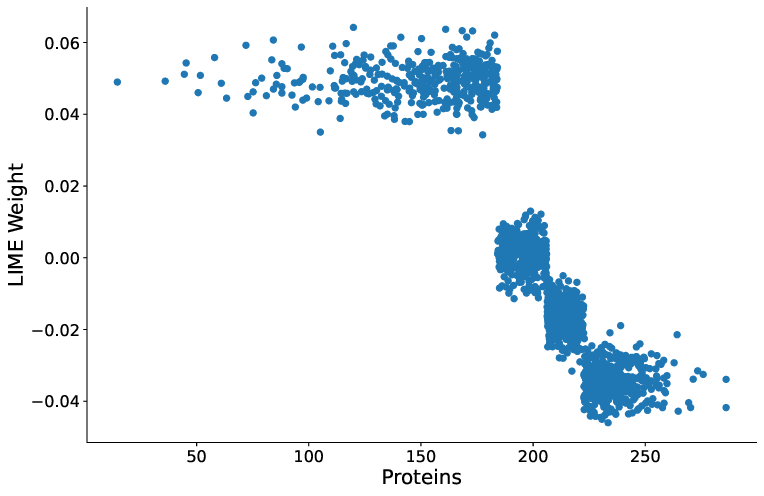}
  \caption{Proteins}
  \label{fig:lime_protein_trend}
\end{subfigure}
\newline
\begin{subfigure}{.5\textwidth}
  \includegraphics[width=0.9\linewidth]{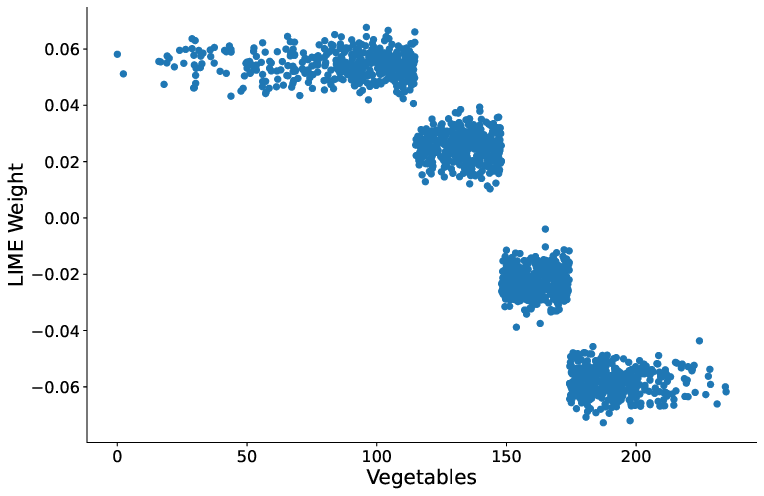}  
  \caption{Vegetables}
  \label{fig:lime_vegetable_trend}
  \end{subfigure}
  \begin{subfigure}{.5\textwidth}
  \includegraphics[width=0.9\linewidth]{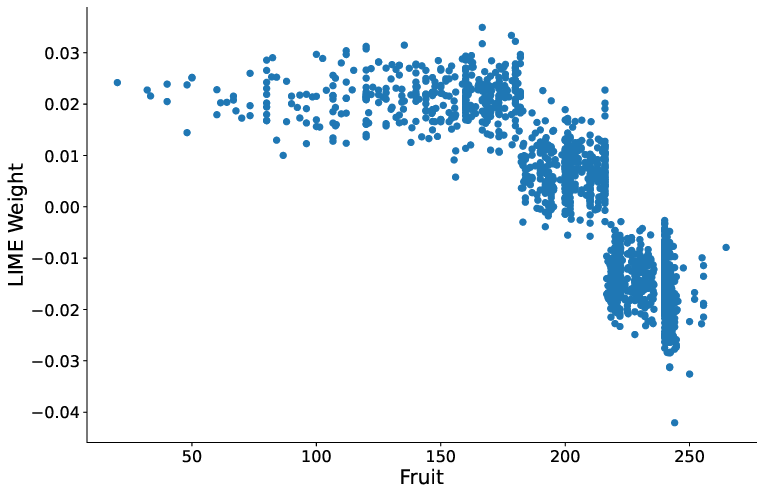}
\caption{Fruit}
\label{fig:lime_fruit_trend}
\end{subfigure}
\caption{LIME nutritional feature trends for RF model.}
\label{fig:LIME_nutritional_feature_trends}
\end{figure}

\end{document}